\documentclass[10pt,twocolumn,letterpaper]{article}

\usepackage{iccv}
\usepackage{times}
\usepackage{epsfig}
\usepackage{graphicx}
\usepackage{amsmath}
\usepackage{amssymb}
\usepackage{subcaption}
\usepackage{enumitem}

\usepackage[pagebackref=true,breaklinks=true,letterpaper=true,colorlinks,bookmarks=false]{hyperref}

 \iccvfinalcopy 

\def\iccvPaperID{717} 

\ificcvfinal\fi
\begin{document}

\title{Generating High-Quality Crowd Density Maps using Contextual Pyramid CNNs}

\author{Vishwanath A. Sindagi and Vishal M. Patel\\
Rutgers University, Department of Electrical and Computer Engineering\\
94 Brett Road, Piscataway, NJ 08854, USA\\
{\tt\small  vishwanath.sindagi@rutgers.edu, vishal.m.patel@rutgers.edu}
}

\maketitle
\thispagestyle{empty}

\begin{abstract}
We present a novel method called Contextual Pyramid CNN (CP-CNN) for generating high-quality crowd density and count estimation by explicitly incorporating global and local contextual information of crowd images. The proposed CP-CNN consists of four modules: Global Context Estimator (GCE), Local Context Estimator (LCE), Density Map Estimator (DME) and a Fusion-CNN (F-CNN). GCE is a VGG-16 based CNN that encodes global context and it is trained to classify input images into different density classes, whereas LCE is another CNN that encodes local context information and it is trained to perform patch-wise classification of input images into different density classes. DME is a multi-column architecture-based CNN that aims to generate high-dimensional feature maps from the input image which are fused with the contextual information estimated by GCE and LCE using F-CNN. To generate high resolution and high-quality density maps, F-CNN uses a set of convolutional and fractionally-strided convolutional layers and it is trained along with the DME in an end-to-end fashion using a combination of adversarial loss and pixel-level Euclidean loss. Extensive experiments on highly challenging datasets show that the proposed method achieves significant improvements over the state-of-the-art methods.

\end{abstract}

\section{Introduction}
\label{sec:intro}
With ubiquitous usage of surveillance cameras and advances in computer vision, crowd scene analysis \cite{li2015crowded,zhan2008crowd} has gained a lot of interest in the recent years. 
In this paper, we focus on the task of estimating crowd count and high-quality density maps which has wide applications in video surveillance \cite{kang2017beyond,xiong2017spatiotemporal}, traffic monitoring, public safety, urban planning \cite{zhan2008crowd}, scene understanding and flow monitoring. Also, the methods developed for crowd counting can be extended to counting tasks in other fields such as cell microscopy \cite{wang2016fast,walach2016learning,lempitsky2010learning,chen2012feature}, vehicle counting \cite{onoro2016towards,zhang2017fcnrlstm,zhang2017understanding,Hsieh_2017_ICCV,toropov2015traffic}, environmental survey \cite{french2015convolutional,zhan2008crowd}, etc. 
\begin{figure}[t!]
\begin{center}
\includegraphics[width=0.48\linewidth]{figures/density/im_25.jpg}
\includegraphics[width=0.48\linewidth]{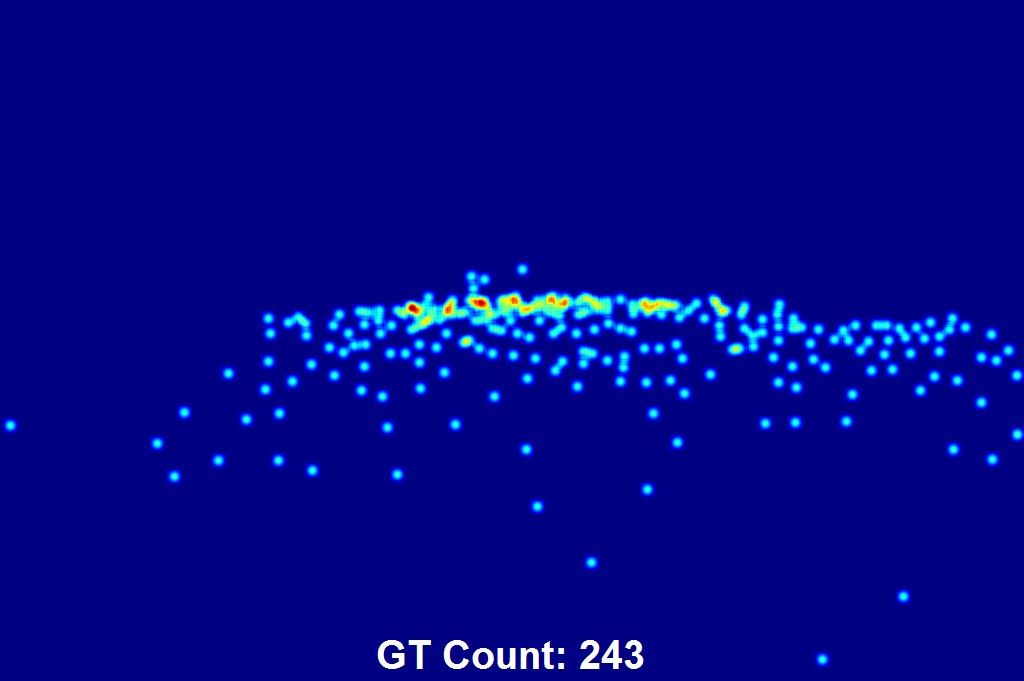}
\includegraphics[width=0.48\linewidth]{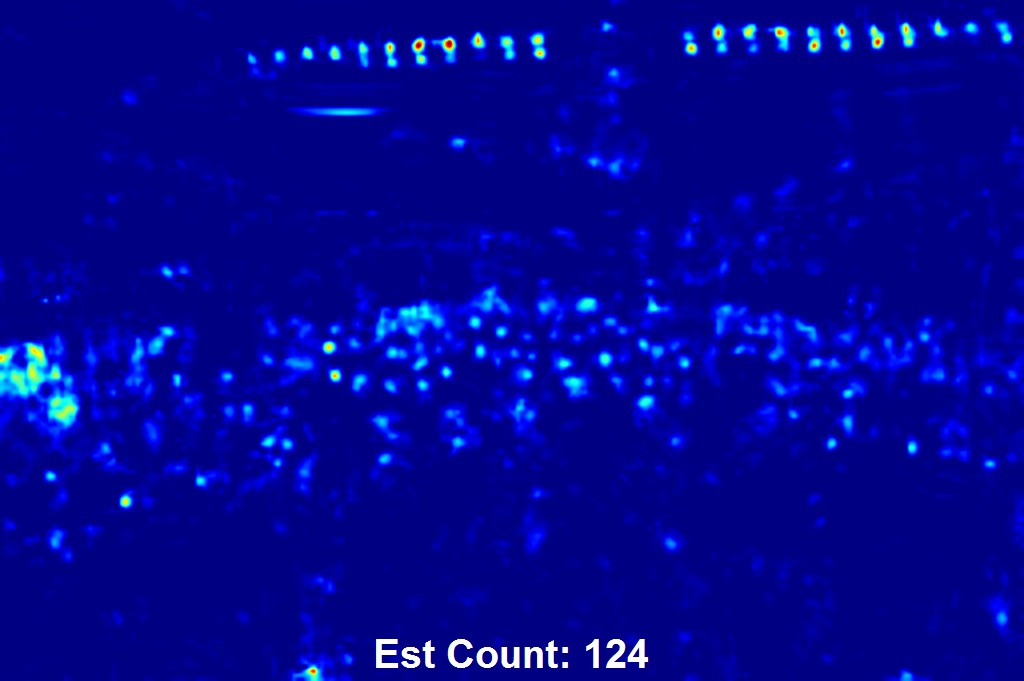}
\includegraphics[width=0.48\linewidth]{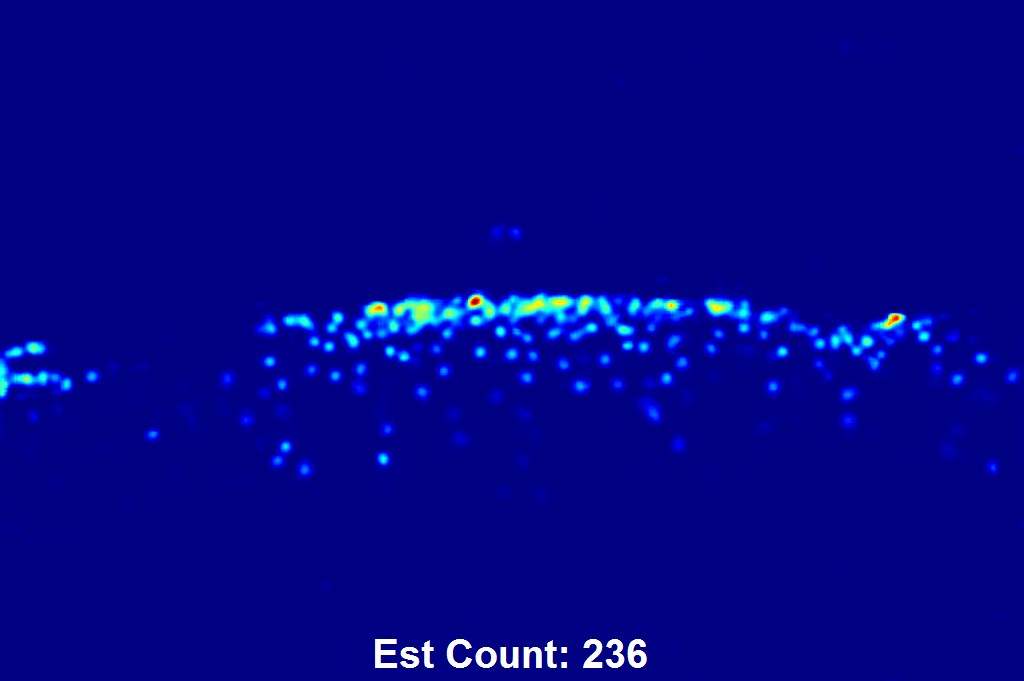}
\end{center}
\vskip -16pt \caption{Density estimation results. Top Left: Input image (from the ShanghaiTech dataset \cite{zhang2016single}). Top Right: Ground truth. Bottom Left: Zhang \etal \cite{zhang2016single} (PSNR: 22.7 dB SSIM: 0.68). Bottom Right: CP-CNN (PSNR: 26.8 dB SSIM: 0.91).}
\label{fig:firstcompare}
\end{figure}
The task of crowd counting and density estimation has seen a significant progress in the recent years. However, due to the presence of various complexities such as occlusions, high clutter, non-uniform distribution of people, non-uniform illumination, intra-scene and inter-scene variations in appearance, scale and perspective, the resulting accuracies are far from optimal.

\begin{figure}[t!]
	\begin{center}
		\includegraphics[width=0.88\linewidth]{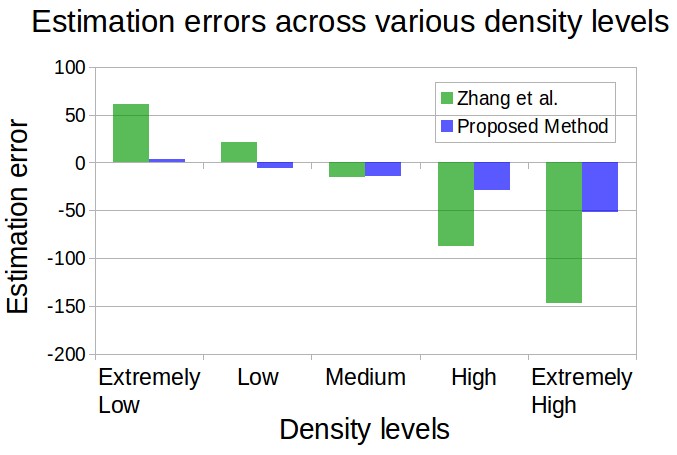}
	\end{center}
	\vskip -17pt \caption{Average estimation errors across various density levels. Current state-of-the-art method \cite{zhang2016single} overestimates/underestimates count in the presence of low-density/high-density crowd.}
	\label{fig:estimation_error_chart}
\end{figure}

Recent CNN-based methods using different multi-scale architectures \cite{zhang2016single,onoro2016towards,sam2017switching} have achieved significant success in addressing some of the above issues, especially in the high-density complex crowded scenes. However, these methods tend to under-estimate or over-estimate count in the presence of high-density and low-density crowd images, respectively (as shown in Fig. \ref{fig:estimation_error_chart}). A potential solution is to use contextual information during the learning process. Several recent works for semantic segmentation \cite{mottaghi2014role}, scene parsing \cite{zhao2016pyramid} and visual saliency \cite{zhao2015saliency} have demonstrated that incorporating contextual information can provide significant improvements in the results. Motivated by their success, we believe that availability of global context shall aid the learning process and help us achieve better count estimation. In addition, existing approaches employ max-pooling layers to achieve minor translation invariance resulting in low-resolution and hence low-quality density maps. Also, to the best of our knowledge, most existing methods concentrate only on the quality of count rather than that of density map. Considering these observations, we propose to incorporate global context into the learning process while improving the quality of density maps. 

To incorporate global context, a CNN-based Global Context Estimator (\textit{GCE}) is trained to encode the context of an input image that is eventually used to aid the density map  estimation process. \textit{GCE} is a CNN-based on VGG-16 architecture. A Density Map Estimator (\textit{DME}), which is a multi-column architecture-based CNN with appropriate max-pooling layers, is used to transform the image into high-dimensional feature maps. Furthermore, we believe that use of local context in the image will guide the \textit{DME} to estimate better quality maps. To this effect, a Local Context Estimator CNN (\textit{LCE}) is trained on input image patches to encode local context information. Finally, the contextual information obtained by LCE and GCE is combined with the output of DME using a Fusion-CNN (\textit{F-CNN}). Noting that the use of max-pooling layers  in \textit{DME} results in low-resolution density maps,  \textit{F-CNN} is constructed using a set of fractionally-strided convolutions \cite{noh2015learning} to increase the output resolution, thereby generating high-quality maps. In a further attempt to improve the quality of density maps, the \textit{F-CNN} is trained using a weighted combination of pixel-wise Euclidean loss and adversarial loss \cite{GAN}. The use of adversarial loss helps us combat the widely acknowledge issue of blurred results obtained by minimizing only the Euclidean loss \cite{GAN_pix2pix2016}. 

The proposed method uses CNN networks to estimate context at various levels for achieving lower count error and better quality density maps. It can be considered as a set of CNNs to estimate pyramid of contexts, hence, the proposed method is dubbed as Contextual Pyramid CNN (CP-CNN). 

To summarize, the following are our main contributions:
\begin{itemize}[topsep=0pt,noitemsep]
  \item We propose a novel Contextual Pyramid CNN (CP-CNN) for crowd count and density estimation that encodes local and global context into the density estimation process.
  \item To the best of our knowledge, ours is the first attempt to concentrate on generating high-quality density maps.  Also, in contrast to the existing methods, we evaluate the quality of density maps generated by the proposed method using different quality measures such as PSNR/SSIM and report state-of-the-art results. 
  \item We use adversarial loss in addition to Euclidean loss for the purpose of crowd density estimation. 
  \item Extensive experiments are conducted on three highly challenging datasets (\cite{zhang2016single,zhang2015cross,idrees2013multi}) and comparisons are performed against several recent state-of-the-art approaches. Further, an ablation study is conducted to demonstrate the improvements obtained by including contextual information and adversarial loss. 
\end{itemize}



\section{Related work}
\label{sec:relatedwork}
Various approaches have been proposed to tackle the problem of crowd counting in images \cite{idrees2013multi,chen2013cumulative,lempitsky2010learning,zhang2015cross,zhang2016single} and videos \cite{brostow2006unsupervised,ge2009marked,rodriguez2011density,chen2015person}.  Initial research focussed on detection style \cite{li2008estimating} and segmentation framework \cite{tu2008unified}. These methods were adversely affected by the presence of occlusions and high clutter in the background. Recent approaches can be broadly categorized into regression-based, density estimation-based and CNN-based methods. We briefly review various methods among these cateogries as follows:\\
\noindent {\bf{Regression-based approaches.}}  To overcome the issues of occlusion and high background clutter, researchers attempted to count by regression where they learn a mapping between features extracted from local image patches to their counts  \cite{chan2009bayesian,ryan2009crowd,chen2012feature}. These methods have two major components: low-level feature extraction and regression modeling. Using a similar approach, Idrees \etal \cite{idrees2013multi} fused count from multiple sources such as head detections, texture elements and frequency domain analysis.\\ 
\noindent{\bf{Density estimation-based approaches.}}  While regression-based approaches were successful in addressing the issues of occlusion and clutter, they ignored important spatial information as they were regressing on the global count. Lempitsky \etal \cite{lempitsky2010learning} introduced a new approach of learning a linear mapping between local patch features and corresponding object density maps using regression. Observing that it is difficult to learn a linear mapping, Pham \etal in \cite{pham2015count} proposed to learn a non-linear mapping between  local patch features and density maps using a random forest framework. Many recent approaches have proposed methods based on density map regression \cite{wang2016fast,xu2016crowd,xia2016block}. A more comprehensive survey of different crowd counting methods can be found in \cite{sindagi2017survey,chen2012feature,li2015crowded,saleh2015recent}.\\
\noindent {\bf{CNN-based methods.}}  Recent success of CNN-based methods in classification and recognition tasks has inspired researchers to employ them for the purpose of crowd counting and density estimation \cite{wang2015deep,zhang2015cross,walach2016learning,skaug2016end}. Walach \etal \cite{walach2016learning} used CNNs with layered training approach. In contrast to the existing patch-based estimation methods, Shang \etal \cite{skaug2016end} proposed an end-to-end estimation method using CNNs by simultaneously learning local and global count on the whole sized input images. Zhang \etal \cite{zhang2016single} proposed a multi-column architecture to extract features at different scales. Similarly, Onoro-Rubio  and L{\'o}pez-Sastre in \cite{onoro2016towards} addressed the scale issue by proposing a scale-aware counting model called Hydra CNN to estimate the object density maps. Boominathan \etal in \cite{boominathan2016crowdnet} proposed to tackle the issue of scale variation using a combination of shallow and deep networks along with an extensive data augmentation by sampling patches from multi-scale image representations. Marsden \etal explored fully convolutional networks \cite{marsden2016fully} and multi-task learning \cite{marsden2017resnetcrowd} for the purpose of crowd counting. 

Inspired by cascaded multi-task learning \cite{ranjan2016hyperface,chen2016cascaded}, Sindagi \etal \cite{sindagi2017cnnbased} proposed to learn a high-level prior and perform density estimation in a cascaded setting.  In contrast to \cite{sindagi2017cnnbased}, the work in this paper is specifically aimed at reducing overestimation/underestimation of count error by systemically leveraging context in the form of crowd density levels at various levels using different networks. Additionally, we incorporate several elements such as local context and adversarial loss aimed at improving the quality of density maps. Most recently, Sam \etal \cite{sam2017switching} proposed a Switching-CNN network that intelligently chooses the most optimal regressor among several independent regressors for a particular input patch.  A comprehensive survey of recent cnn-based methods for crowd counting can be found in \cite{sindagi2017survey}.
\begin{figure}[t]
	\begin{center}
		\includegraphics[width=.9\linewidth]{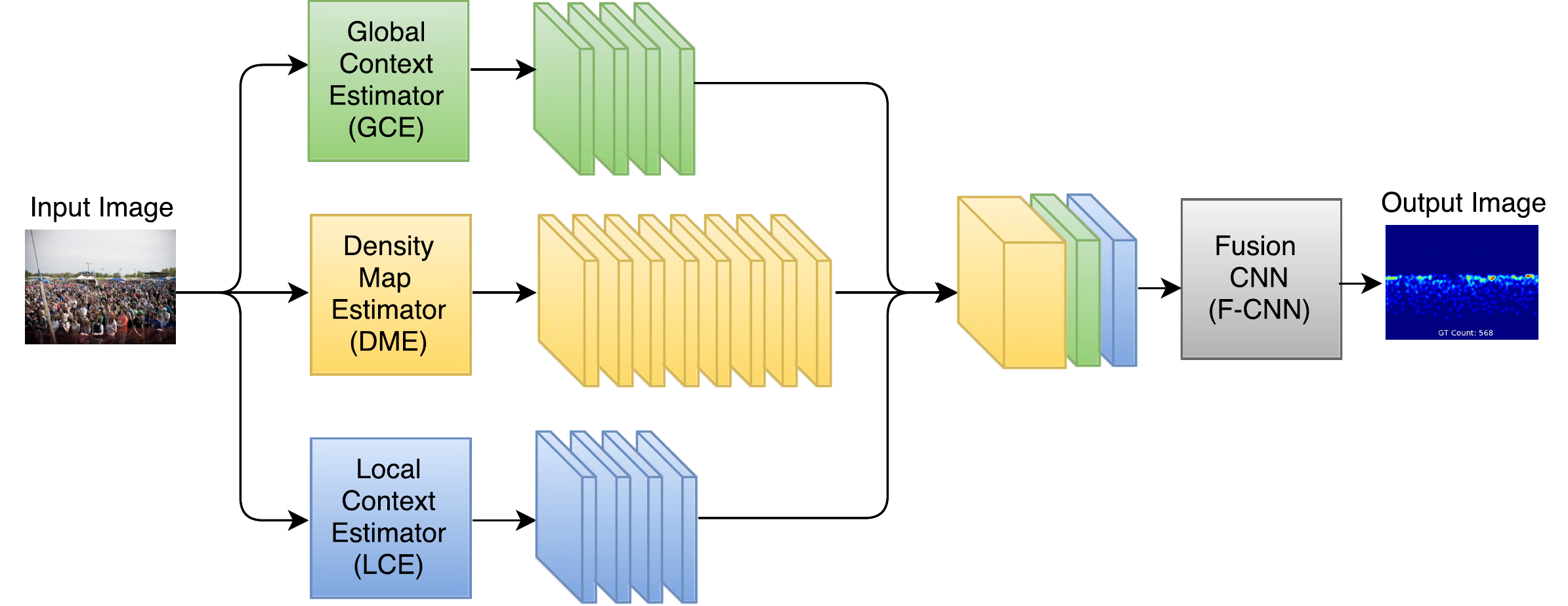}
	\end{center}
	\vskip -16pt \caption{Overview of the proposed CP-CNN architecture. The network incorporates global and local context using \textit{GCE} and \textit{LCE} respectively. The context maps are concatenated with the output of \textit{DME} and further processed by \textit{F-CNN} to estimate high-quality density maps.}
	\label{fig:arch}
\end{figure}
Recent works using multi-scale and multi-column architectures \cite{zhang2016single,onoro2016towards,walach2016learning} have demonstrated considerable success in achieving lower count errors. We make the following observations regarding these recent state-of-the-art approaches: 1. These methods do not explicitly incorporate contextual information which is essential for achieving further improvements. 2. Though existing approaches regress on density maps, they are more focussed on improving count errors rather than quality of the density maps, and 3. Existing CNN-based approaches are trained using a pixel-wise Euclidean loss which results in blurred density maps. In view of these observations, we propose a novel method to learn global and local contextual information from images for achieving better count estimates and high-quality density maps. Furthermore, we train the CNNs in a Generative Adversarial Network (GAN) based framework \cite{GAN}  to exploit the recent success of adversarial loss to achieve high-quality and sharper density maps.

\section{Proposed method (CP-CNN)}
\label{sec:proposedmethod}

The proposed CP-CNN method consists of a pyramid of context estimators and a Fusion-CNN as illustrated in Fig. \ref{fig:arch}. It consists of four modules: \textit{GCE},  \textit{LCE},  \textit{DME},  and  \textit{F-CNN}. \textit{GCE} and \textit{LCE} are CNN-based networks that encode global and local context present in the input image respectively. \textit{DME} is a multi-column CNN that performs the initial task of transforming the input image to high-dimensional feature maps. Finally, \textit{F-CNN} combines contextual information from \textit{GCE} and \textit{LCE} with high-dimensional feature maps from \textit{DME} to produce high-resolution and high-quality density maps. These modules are discussed in detail as follows.

\subsection{Global Context Estimator (GCE)}

As discussed in Section ~\ref{sec:intro}, though recent state-of-the-art multi-column or multi-scale methods \cite{zhang2016single,onoro2016towards,walach2016learning} achieve significant improvements in the task of crowd count estimation, they either underestimate or overestimate  counts in high-density and low-density crowd images respectively (as explained in Fig. ~\ref{fig:estimation_error_chart}). We believe it is important to explicilty model context present in the image to reduce the estimation error. To this end, we associate global context with the level of density present in the image by considering the task of learning global context as classifying the input image into five different classes: extremely low-density (ex-lo), low-density (lo), medium-density (med), high-density (hi) and extremely high-density (ex-hi). Note that the number of classes required is dependent on the crowd density variation in the dataset. A dataset containing large variations may require higher number of classes. In our experiments, we obtained significant improvements using five categories of density levels. 

\begin{figure}[htp!]
\begin{center}
\includegraphics[width=0.85\linewidth]{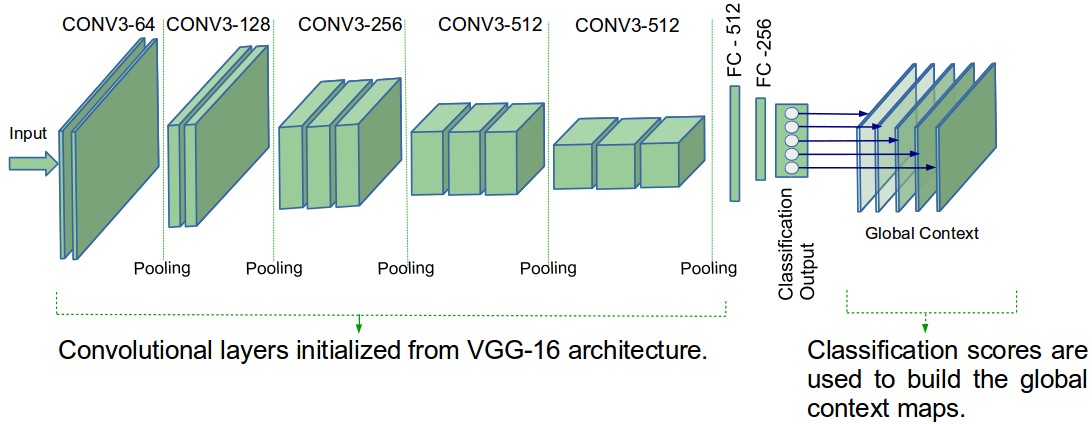}
\end{center}
  \vskip -17pt \caption{Global context estimator based on VGG-16 architecture. The network is trained to classify the input images into various density levels thereby encoding the global context present in the image.}
\label{fig:gce}
\end{figure}

In order to learn the classification task, a VGG-16 \cite{simonyan2014very} based network is fine-tuned with the crowd training data. Network used for \textit{GCE} is as shown in Fig. \ref{fig:gce}. The convolutional layers from the VGG-16 network are retained, however, the last three fully connected layers are replaced with a different configuration of fully connected layers in order to cater to our task of classification into five categories. Weights of the last two convolutional layers are fine-tuned while keeping the weights fixed for the earlier layers. The use of pre-trained VGG network results in faster convergence as well as better performance in terms of context estimation.

\subsection{Local Context Estimator (LCE)}
\begin{figure}[htp!]
\begin{center}
\includegraphics[width=0.65\linewidth]{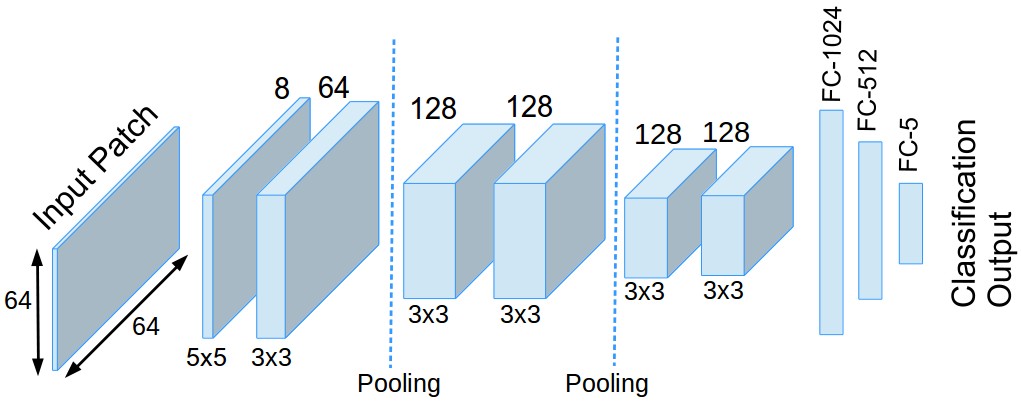}
\end{center}
  \vskip -18pt \caption{Local context estimator: The network is trained to classify local input patches into various density levels thereby encoding the local context present in the image.}
\label{fig:lce}
\end{figure}

Existing methods for crowd density estimation have primarily focussed on achieving lower count errors rather than estimating better quality density maps. As a result, these methods produce low-quality density maps as shown in Fig. \ref{fig:firstcompare}. After an analysis of these results, we believe that some kind of local contextual information can aid us to achieve better quality maps. To this effect, similar to \textit{GCE}, we propose to learn an image's local context by learning to classify it's local patches into one of the five classes: \{ex-lo, lo, med, hi, ex-hi\}. The local context is learned by the \textit{LCE} whose architecture shown in Fig. \ref{fig:lce}. It is composed of a set of convolutional and max-pooling layers followed by 3 fully connected layers with appropriate drop-out layers after the first two fully connected layers. Every convolutional and fully connected layer is followed by a ReLU layer except for the last fully connected layer which is followed by a sigmoid layer. 

\subsection{Density Map Estimator (DME)}
\label{ssec:dme}
\begin{figure}[htp!]
\begin{center}
\includegraphics[width=0.7\linewidth]{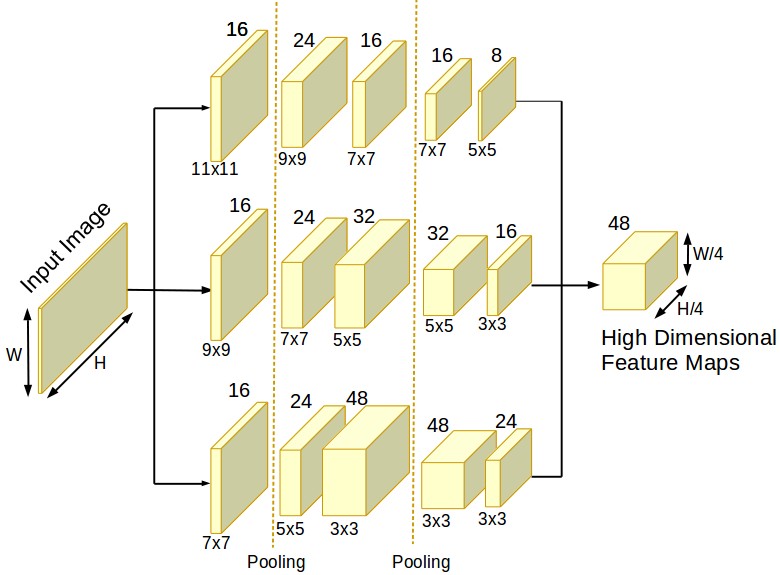}
\end{center}
  \vskip -18pt \caption{Density Map Estimator: Inspired by Zhang \etal \cite{zhang2016single}, DME is a multi-column architecture. In contrast to \cite{zhang2016single}, we use slightly deeper columns with different number of filters and filter sizes.}
\label{fig:dme}
\end{figure}

The aim of \textit{DME} is to transform the input image into a set of high-dimensional feature maps which will be concatenated with the contextual information provided by \textit{GCE} and \textit{LCE}. Estimating density maps from high-density crowd images is especially challenging due to the presence of heads with varying sizes in and across images. Previous works on multi-scale \cite{onoro2016towards} or multi-column \cite{zhang2016single} architectures have demonstrated abilities to handle the presence of considerably large variations in object sizes by achieving significant improvements in such scenarios. Inspired by the success of these methods, we use a multi-column architecture similar to \cite{zhang2016single}. However, notable differences compared to their work are that our columns are much deeper and have different number of filters and filter sizes that are optimized for lower count estimation error. Also, in this work, the multi-column architecture is used to transform the input into a set of high-dimensional feature map rather than using them directly to estimate the density map. Network details for \textit{DME} are illustrated in Fig. \ref{fig:dme}. 

It may be argued that since the \textit{DME} has a pyramid of filter sizes, one may be able to increase the filter sizes and number of columns to address larger variation in scales. However, note that addition of more columns and the filter sizes will have to be decided based on the scale variation present in the dataset, resulting in new network designs that cater to different datasets containing different scale variations. Additionally, deciding the filter sizes will require time consuming experiments. With our network, the design remains consistent across all datasets, as the context estimators can be considered to perform the task of coarse crowd counting.

\subsection{Fusion-CNN (F-CNN)}
The contextual information from \textit{GCE} and \textit{LCE} are combined with the high-dimensional feature maps from \textit{DME} using \textit{F-CNN}.  The \textit{F-CNN} automatically learns to incorporate the contextual information estimated by context estimators. The presence of max-pooling layers in the \textit{DME} network (which are essential to achieve translation invariance) results in down-sampled feature maps and loss of details. Since, the aim of this work is to estimate high-resolution and high-quality density maps, \textit{F-CNN} is constructed using a set of convolutional and fractionally-strided convolutional layers. The set of fractionally-strided convolutional layers help us to restore details in the output density maps. The following structure is used for \textit{F-CNN}: \emph{CR(64,9)-CR(32,7)-TR(32)-CR(16,5)-TR(16)-C(1,1)},
where, $C$ is convolutional layer, $R$ is ReLU layer, $T$ is fractionally-strided convolution layer and the first number inside every brace indicates the number of filters while the second number indicates filter size. Every fractionally-strided convolution layer increases the input resolution by a factor of 2, thereby ensuring that the output resolution is the same as that of input. 

Once the context estimators are trained, \textit{DME} and \textit{F-CNN} are trained in an end-to-end fashion. Existing methods for crowd density estimation use Euclidean loss to train their networks. It has been widely acknowledged that minimization of $L_2$ error results in blurred results especially for image reconstruction tasks \cite{GAN_pix2pix2016,johnson2016perceptual,zhang17ICCV,zhang2017image,zhang2017joint}. Motivated by these observations and the recent success of GANs for overcoming the issues of L2-minimization \cite{GAN_pix2pix2016}, we attempt to further improve the quality of density maps by minimizing a weighted combination of pixel-wise Euclidean loss and adversarial loss. The loss  for training \textit{F-CNN} and \textit{DME} is defined as follows:
\begin{gather}
\label{eq:Eu_ADV}L_T = L_E + \lambda_a L_A,  \\
L_E= \frac{1}{WH}\sum_{w=1}^{W}\sum_{h=1}^{H} \|\phi({X^{w,h}})-(Y^{w,h})\|_2,  \\
L_{A}= -\log(\phi_D(\phi(X)),
\end{gather}
where, $L_T$ is the overall loss, $L_E$ is the pixel-wise Euclidean loss between estimated density map and it's corresponding ground truth, $\lambda_a$ is a weighting factor, $L_{A}$ is the adversarial loss, $X$ is the input image of dimensions $W\times H$, $Y$ is the ground truth density map, $\phi$ is the network consisting of \textit{DME} and \textit{F-CNN} and $\phi_D$ is the discriminator sub-network for calculating the adversarial loss. Following structure is used for the discriminator sub-network: \emph{CP(64)-CP(128)-M-CP(256)-M-CP(256)-CP(256)-M-C(1)-Sigmoid},
where $C$ represents convolutional layer, $P$ represents PReLU layer and $M$ is max-pooling layer. 

\section{Training and evaluation details}
In this section, we discuss details of the training and evaluation procedures. 

\noindent \textbf{Training details:} Let $D$ be the original training dataset. Patches $1/4^{th}$ the size of original image are cropped from 100 random locations from every image in $D$. Other augmentation techniques like horizontal flipping and noise addition are used to create another 200 patches. The random cropping and augmentation resulted in a total of 300 patches per image in the training dataset. Let this set of images be called as $D_{dme}$. Another training set $D_{lc}$ is formed by cropping patches of size $64\times 64$ from 100 random locations in every training image in $D$. 

\textit{GCE} is trained using the dataset  $D_{dme}$. The corresponding ground truth categories for each image is determined based on the number of people present in it. Note that the images are resized to $224\times 224$ before feeding them into the VGG-based \textit{GCE} network. The network is then trained using the standard cross-entropy loss. \textit{LCE} is trained using the $64\times 64$ patches in $D_{lc}$. The ground truth categories of the training patches is determined based on the number of people present in them. The network is then trained using the standard cross-entropy loss.

Next, the \textit{DME} and \textit{F-CNN} networks are trained in an end-to-end fashion  using input training images from $D_{dme}$ and their corresponding global and local contexts\footnote{Once \textit{GCE} and \textit{LCE} are trained, their weights are frozen.}. The global context ($F_{gc}^i$) for an input training image $X^i$ is obtained in the following way. First,  an empty global context $F_{gc}^i$ of dimension $5 \times W_i/4 \times H_i/4$ is  created, where $W_i \times H_i$ is the dimension of $X_i$. Next, a set of classification scores $y_{gc}^{i,j} (j=1...5)$ is obtained by feeding $X_i$ to \textit{GCE}. Each feature map in global context $F_{gc}^{i,j}$ is then filled with the corresponding classification score $y_g^{i,j}$. The local context ($F_{lc}^i$) for $X^i$ is obtained in the following way. An empty local context $F_{lc}^i$ of dimension $5 \times W_i \times H_i$ is first created. A sliding window classifier (\textit{LCE}) of size 
$64\times 64$ is run on $X_i$ to obtain the classification score $y_{lc}^{i,j,w} (j=1...5)$ where $w$ is the window location.  The classification scores $y_{lc}^{i,j,w}$ are used to fill the corresponding window location $w$ in the respective local context map $F_{gc}^{i,j}$. $F_{gc}^{i,j}$ is then resized to a size of  $W_i/4 \times H_i/4$. After the context maps are estimated, $X_i$ is fed to \textit{DME} to obtain a high-dimensional feature map $F_{dme}^i$ which is concatenated with $F_{gc}^{i}$ and $F_{lc}^{i}$. These concatenated feature maps are then fed into \textit{F-CNN}. The two CNNs (\textit{DME} and \textit{F-CNN}) are trained in an end-to-end fashion by minimizing the weighted combination of pixel-wise Euclidean loss and adversarial loss (given by \eqref{eq:Eu_ADV}) between the estimated and ground truth density maps. 
\linebreak

\noindent \textbf{Inference details:} Here, we describe the process to estimate the density map of a test image $X_i^t$. First, the global context map $F_{tgc}^i$ for  $X_i^t$  is calculated in the following way. The test image $X_i^t$ is divided into non-overlapping blocks of size $W_{i}^t/4 \times H_{i}^t/4$. All blocks are then fed into \textit{GCE} to obtain their respective classification scores. As in training, the classification scores are used to build the context maps for each block to obtain the final global context feature map $F_{tgc}^i$. Next, the local context map $F_{tlc}^i$ for  $X_i^t$ is calculated in the following way: A sliding window classifier (\textit{LCE}) of size $64\times 64$ is run  across  $X_i^t$ and the classification scores from every window are used to build the local context $F_{tlc}^i$. Once the context information is obtained, $X_i^t$ is fed into \textit{DME} to obtain high-dimensional feature maps  $F_{tdme}^i$. $F_{tdme}^i$ is concatenated with  $F_{tgc}^i$ and $F_{tlc}^i$ and fed into \textit {F-CNN} to obtain the output density map. Note that due to additional context processing, inference using the proposed method is computationally expensive as compared to earlier methods such as \cite{zhang2016single,sam2017switching}.

\section{Experimental results}
\label{sec:results}
In this section, we present the experimental details and evaluation results on three publicly available  datasets. First, the results of an ablation study conducted to demonstrate the effects of each module in the architecture is discussed. Along with the ablation study, we also perform a detailed comparison of the proposed method against a recent state-of-the-art-method \cite{zhang2016single}. This detailed analysis contains comparison of count metrics defined by  \eqref{eq:count_error}, along with qualitative and quantitative comparison of the estimated density maps. The quality of density maps is measured using two standard metrics: PSNR (Peak Signal-to-Noise Ratio) and SSIM (Structural Similarity in Image \cite{SSIM}). The count error is measured using Mean Absolute Error (MAE) and Mean Squared Error (MSE):
\begin{align}
\label{eq:count_error}
 MAE = \frac{1}{N}\sum_{i=1}^{N}|y_i-y'_i|, 
 MSE = \sqrt{\frac{1}{N}\sum_{i=1}^{N}|y_i-y'_i|^2},
\end{align}
where $N$ is number of test samples, $y_i$ is the ground truth count and $y'_i$ is the estimated count corresponding to the $i^{th}$ sample. The ablation study is followed by a discussion and comparison of proposed method's results against several recent state-of-the-art methods on three datasets: ShanghaiTech \cite{zhang2016single}, WorldExpo '10 \cite{zhang2015cross} and UCF\textunderscore CROWD\textunderscore 50 \cite{idrees2013multi}.   

\subsection{Ablation study using ShanghaiTech Part A}
\label{ssec:analysis}

In this section, we perform an ablation study to demonstrate the effects of different modules in the proposed method. Each module is added sequentially to the network and results for each configuration are compared. Following four configurations are evaluated:  (1) \textit{DME}: The high-dimensional feature maps of \textit{DME} are combined using 1$\times$1 conv layer whose output is used to estimate the density map. $L_E$ loss is minimized to train the network. (2) \textit{DME} with only \textit{GCE} and \textit{F-CNN}: The output of \textit{DME} is concatenated with the global context. \textit{DME} and \textit{F-CNN} are trained to estimate the density maps by minimizing $L_E$ loss. (3) \textit{DME} with \textit{GCE}, \textit{LCE} and \textit{F-CNN}. In addition to the third configuration, local context is also used in this case and the network is trained using $L_E$ loss. (4) \textit{DME} with \textit{GCE}, \textit{LCE} and \textit{F-CNN} with $L_{A} + L_E$ (entire network). These results are compared with a fifth configuration: Zhang \textit{et al.} \cite{zhang2016single} (which is a recent state-of-the-art method) in order to gain a perspective of the improvements achieved by the proposed method and its various modules. 

The evaluation is performed on Part A of ShanghaiTech  \cite{zhang2016single} dataset which contains 1198 annotated images with a total of 330,165 people. This dataset consists of two parts: Part A with 482 images and Part B with 716 images. Both parts are further divided into training and test datasets with training set of Part A containing 300 images and that of Part B containing 400 images. Rest of the images are used as test set. Due to the presence of large variations in density, scale and appearance of people across images in the Part A of this dataset, estimating the count with high degree of accuracy is difficult. Hence, this dataset was chosen for the detailed analysis of performance of the proposed architecture.
 
\begin{figure*}[ht!]
\begin{center}
\begin{subfigure}[t]{0.17\textwidth}
\includegraphics[width=\textwidth]{figures/d1/IMG_88.jpg}
\includegraphics[width=\textwidth]{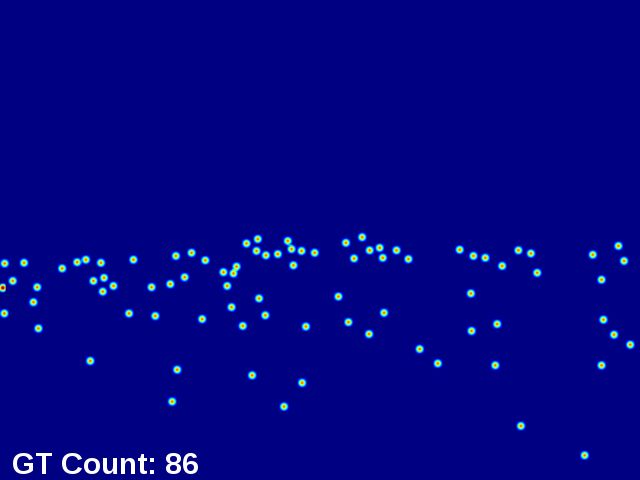}
\includegraphics[width=\linewidth]{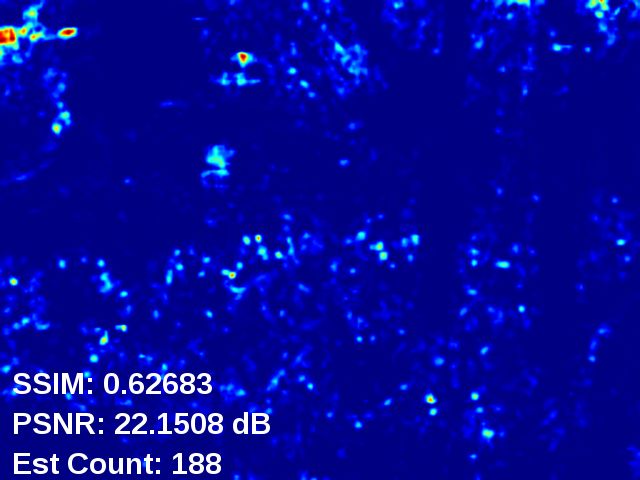}
\includegraphics[width=\linewidth]{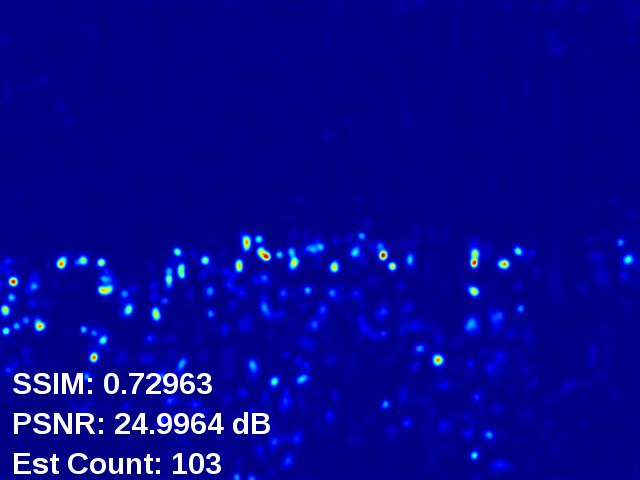}
\includegraphics[width=\linewidth]{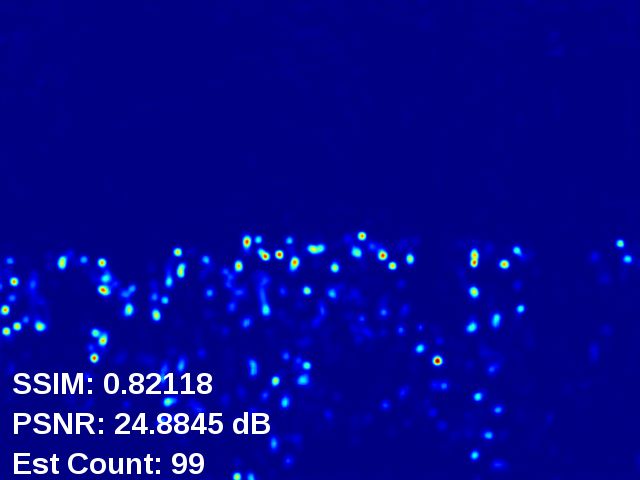}
\includegraphics[width=\linewidth]{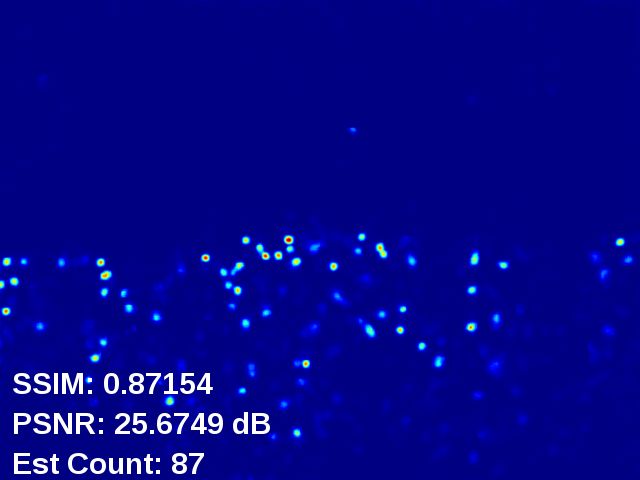}
\end{subfigure}
\begin{subfigure}[t]{0.17\textwidth}
\includegraphics[width=\textwidth]{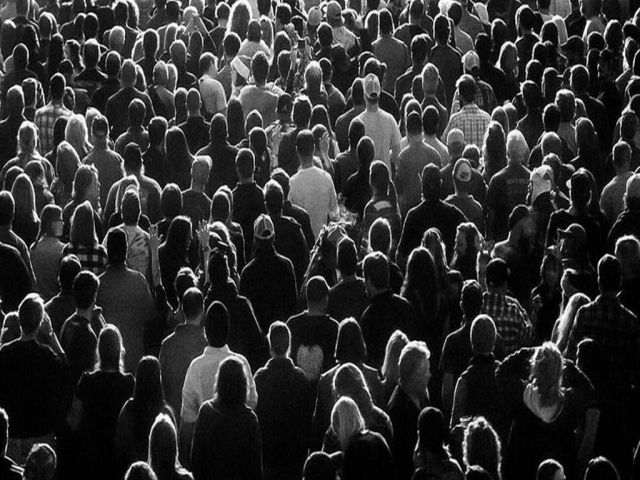}
\includegraphics[width=\textwidth]{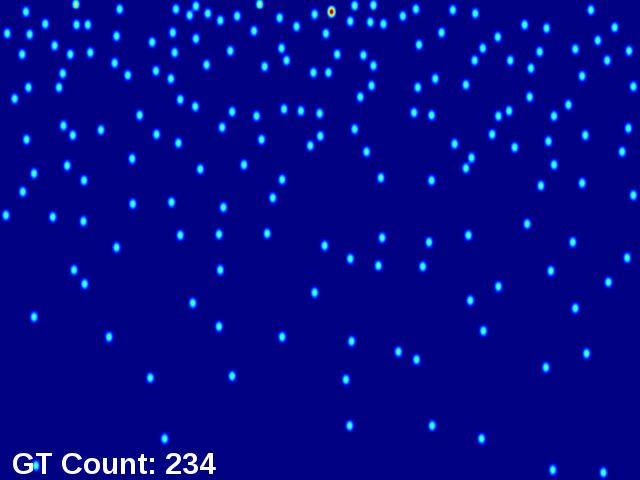}
\includegraphics[width=\linewidth]{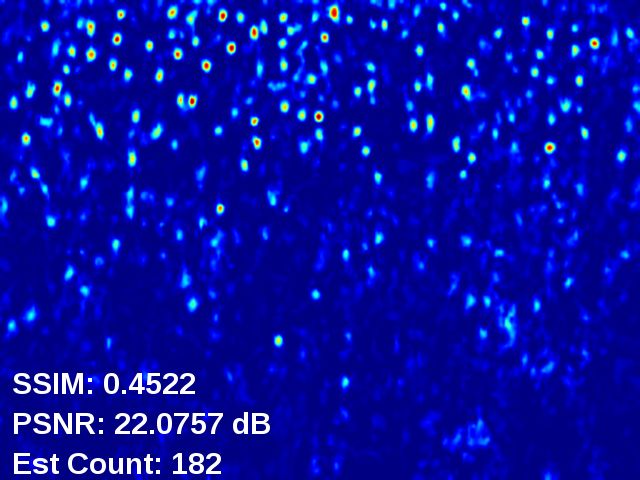}
\includegraphics[width=\linewidth]{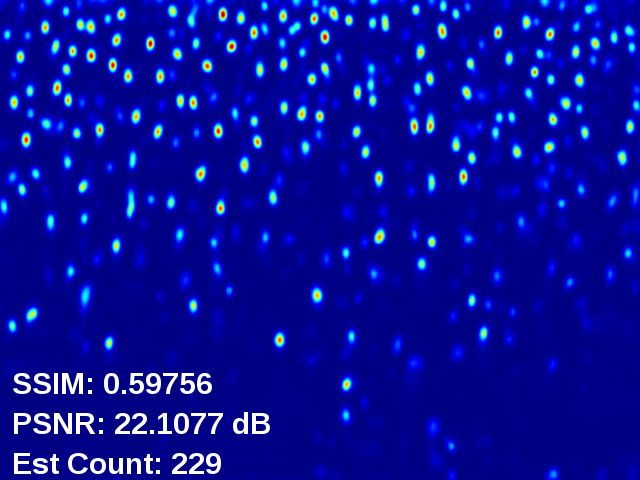}
\includegraphics[width=\linewidth]{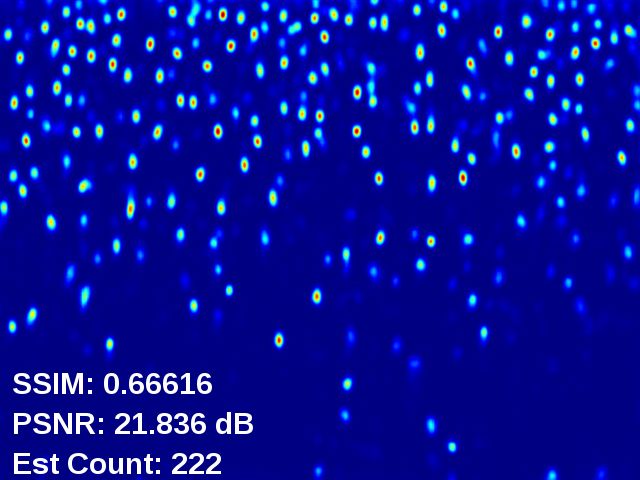}
\includegraphics[width=\linewidth]{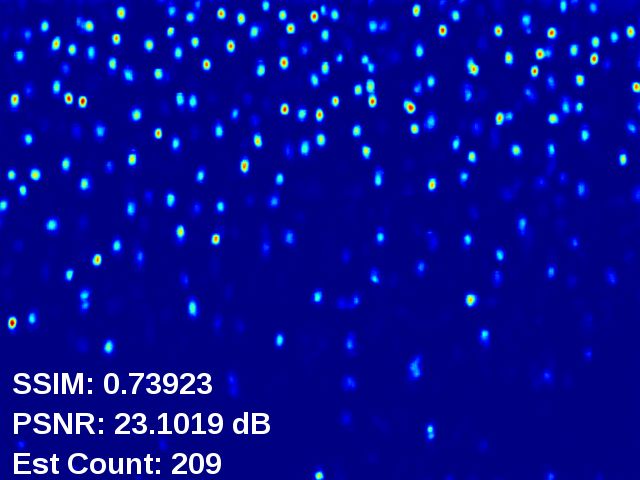}
\end{subfigure}
\begin{subfigure}[t]{0.17\textwidth}
\includegraphics[width=\textwidth]{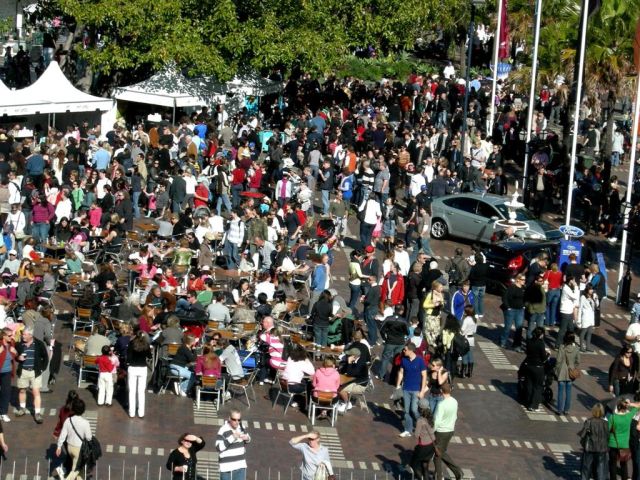}
\includegraphics[width=\textwidth]{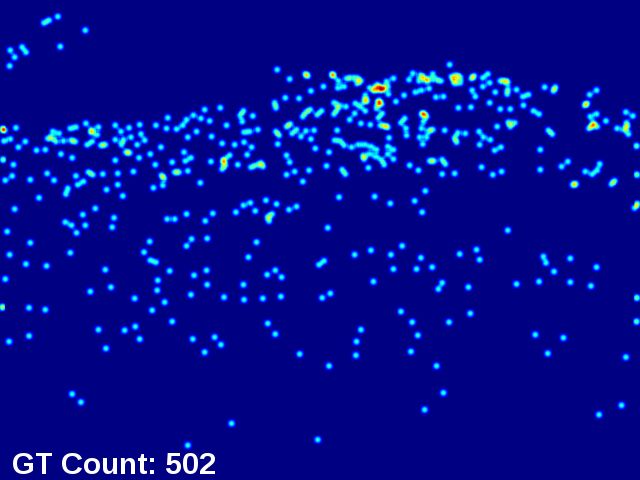}
\includegraphics[width=\linewidth]{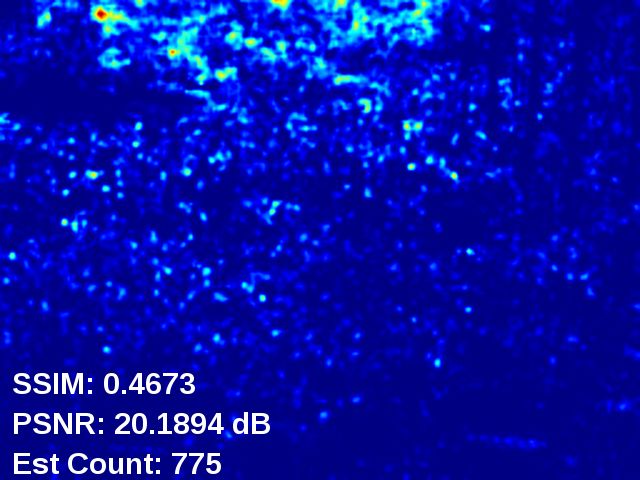}
\includegraphics[width=\linewidth]{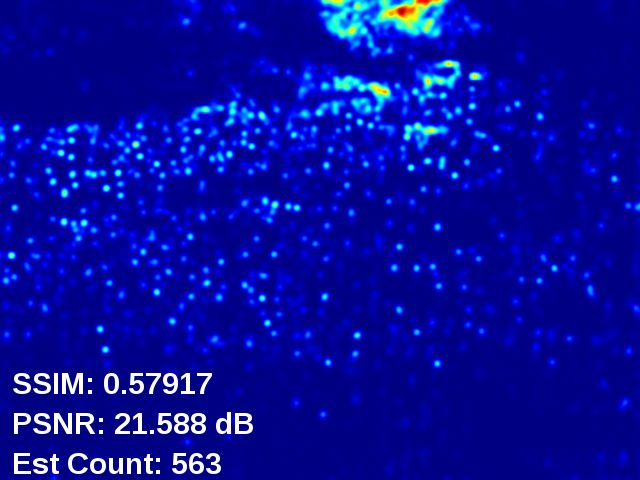}
\includegraphics[width=\linewidth]{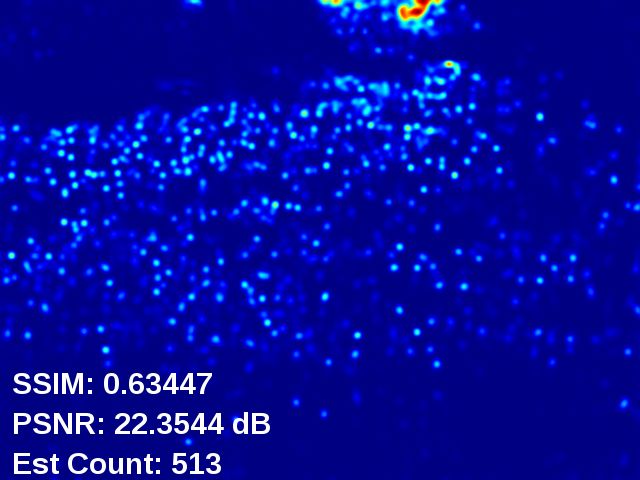}
\includegraphics[width=\linewidth]{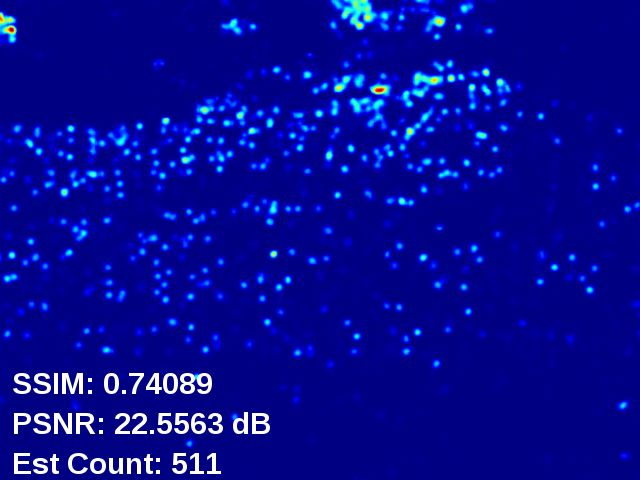}
\end{subfigure}
\begin{subfigure}[t]{0.17\textwidth}
\includegraphics[width=\textwidth]{figures/d1/IMG_21.jpg}
\includegraphics[width=\textwidth]{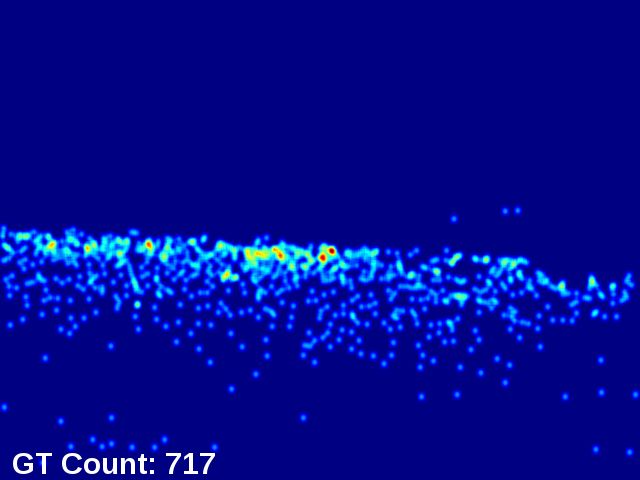}
\includegraphics[width=\linewidth]{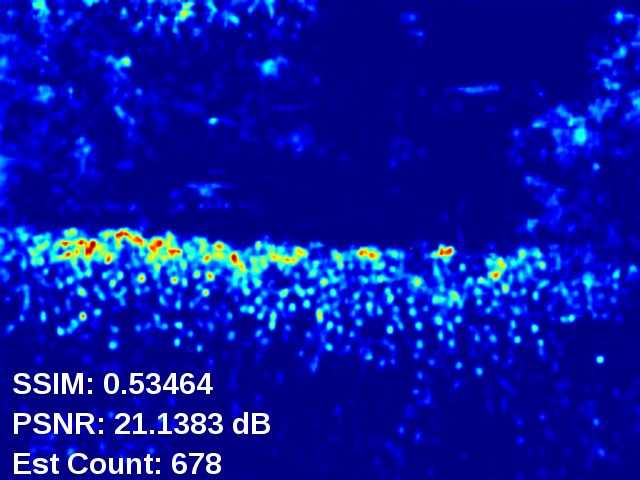}
\includegraphics[width=\linewidth]{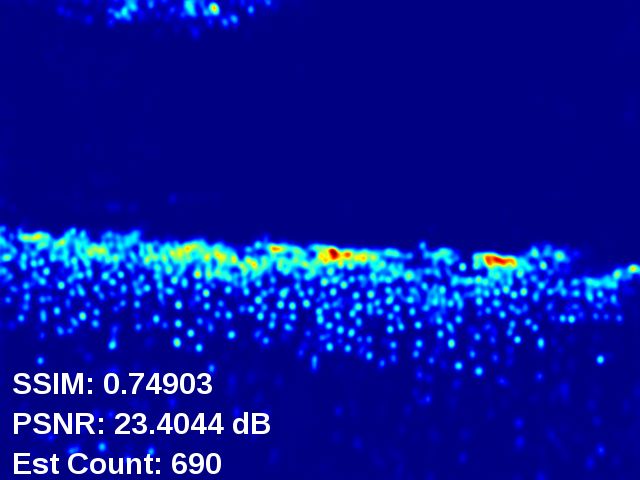}
\includegraphics[width=\linewidth]{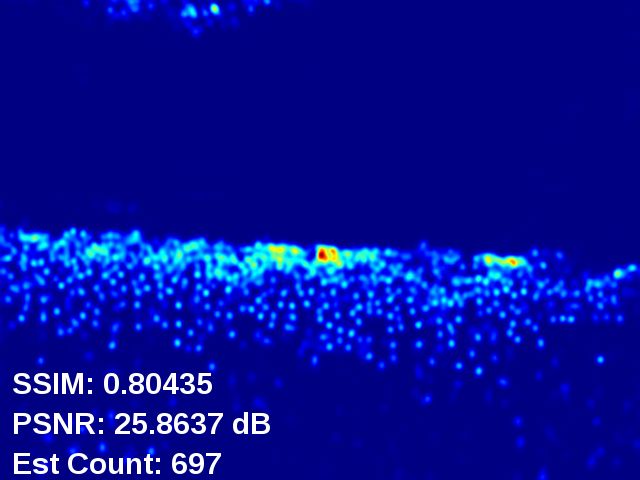}
\includegraphics[width=\linewidth]{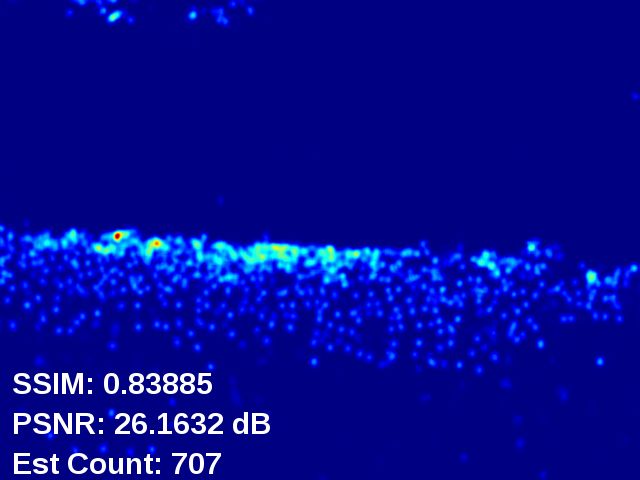}
\end{subfigure}
\begin{subfigure}[t]{0.17\textwidth}
\includegraphics[width=\textwidth]{figures/d1/IMG_63.jpg}
\includegraphics[width=\textwidth]{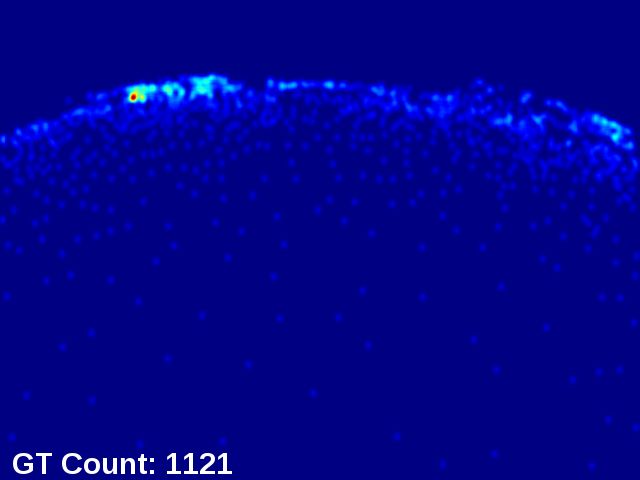}
\includegraphics[width=\linewidth]{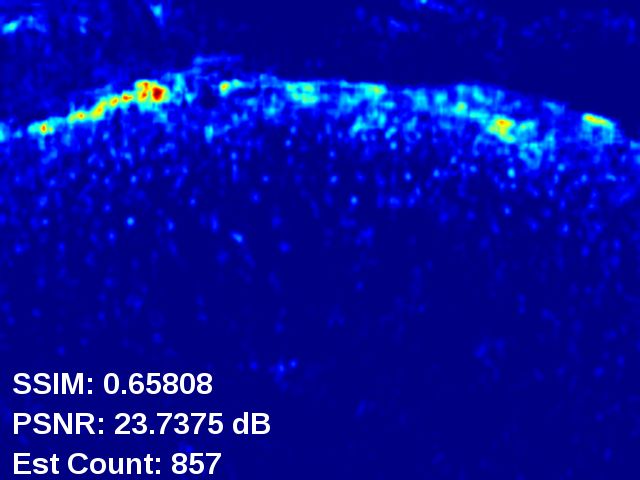}
\includegraphics[width=\linewidth]{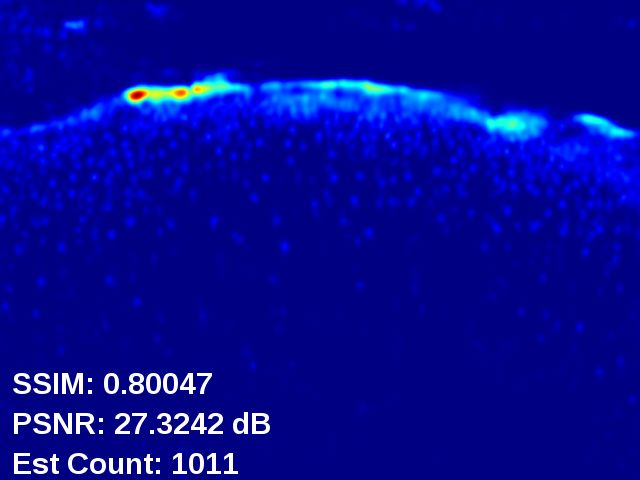}
\includegraphics[width=\linewidth]{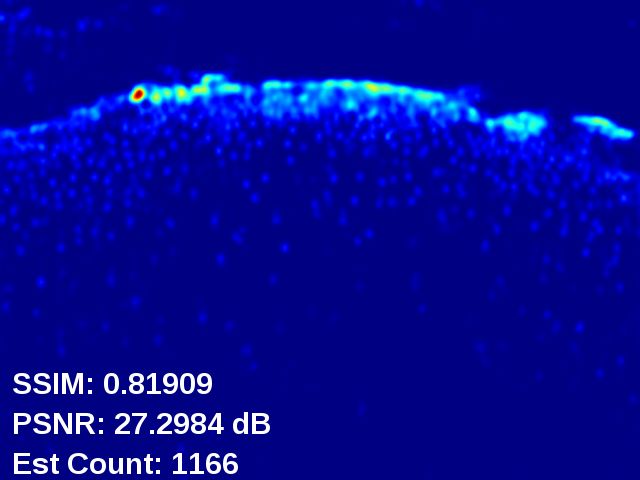}
\includegraphics[width=\linewidth]{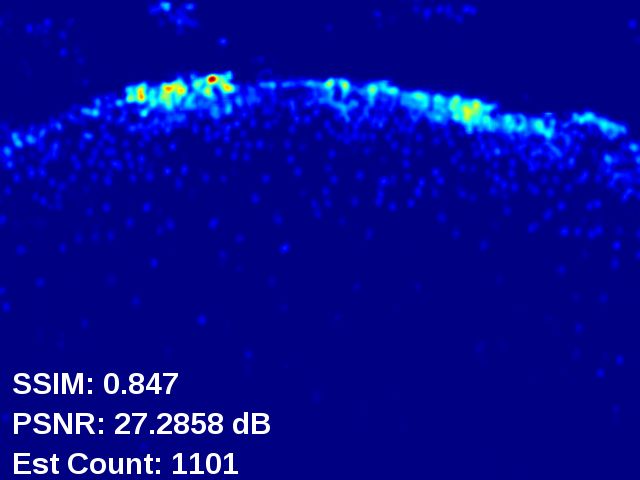}
\end{subfigure}

\end{center}
\vskip -16pt\caption{Comparison of results from different configurations of the proposed network along with Zhang \etal \cite{zhang2016single}. Top Row: Sample input images from the ShanghaiTech dataset. Second Row: Ground truth. Third Row: Zhang \etal \cite{zhang2016single}. (Loss of details can be observed). Fourth Row: \textit{DME}. Fifth Row: \textit{DME} + \textit{GCE} + \textit{F-CNN}. Sixth Row:\textit{DME} + \textit{GCE} + \textit{LCE} + \textit{F-CNN}. Bottom Row: \textit{DME} + \textit{GCE} + \textit{LCE} + \textit{F-CNN}  with adversarial loss. Count estimates and the quality of density maps improve after inclusion of contextual information and adversarial loss.}
\label{fig:detailedanalysis}
\end{figure*}

\begin{table}[t!]
\centering
\resizebox{0.8\linewidth}{!}{
\begin{tabular}{|l|c|c|c|c|}
\hline
                & \multicolumn{2}{c|}{\begin{tabular}[c]{@{}l@{}}Count \\ estimation \\ error\end{tabular}} & \multicolumn{2}{c|}{\begin{tabular}[c]{@{}l@{}}Density map \\quality \end{tabular}} \\ \hline
Method          &  MAE          & MSE          & PSNR           & SSIM           \\ \hline\hline
Zhang \textit{et al.}\cite{zhang2016single}   & 110.2        & 173.2        & 20.91         & 0.52        \\ \hline
\textit{DME}   & 104.3        & 154.2        & 20.92         & 0.54    \\ \hline
\textit{DME}+\textit{GCE}+\textit{FCNN}& 89.9        & 127.9        & 20.97         & 0.61         \\ \hline
\begin{tabular}[c]{@{}l@{}}\textit{DME} + \textit{GCE} +\\ \textit{LCE} + \textit{FCNN}\end{tabular}& {76.1}        & {110.2}        & {21.4}         & {0.65}         \\ \hline
\begin{tabular}[c]{@{}l@{}}\textit{DME}+\textit{GCE}+\textit{LCE}+\\ \textit{FCNN} with $L_{A}$+$L_E$\end{tabular}&\textbf{73.6}        & \textbf{106.4}        & \textbf{21.72}         & \textbf{0.72}        
 \\ \hline
\end{tabular}
}
\vskip-8pt\caption{Estimation errors for different configurations of the proposed network on ShanghaiTech Part A\cite{zhang2016single}. Addition of contextual information and the use of adversarial loss progressively improves the count error and the quality of density maps.}
\label{tab:detailedanalysis}
\end{table}

Count estimation errors and quality metrics of the estimated density images for the various configurations are tabulated in Table \ref{tab:detailedanalysis}. We make the following observations: (1) The network architecture for \textit{DME} used in this work is different from Zhang \etal \cite{zhang2016single} in terms of column depths, number of filters and filter sizes. These changes improve the count estimation error as compared to \cite{zhang2016single}. However, no significant improvements are observed in the quality of density maps. (2) The use of global context in (\textit{DME} + \textit{GCE} + \textit{F-CNN}) greatly reduces the count error from the previous configurations. Also, the use of \textit{F-CNN} (which is composed of fractionally-strided convolutional layers), results in considerable improvement in the quality of density maps. (3) The addition of local context and the use of adversarial loss progressively reduces the count error while achieving better quality in terms of PSNR and SSIM.

Estimated density maps from various configurations on sample input images are shown in Fig. \ref{fig:detailedanalysis}. It can be observed that the density maps generated using Zhang \etal \cite{zhang2016single} and \textit{DME} (which regress on low-resolution maps) suffer from loss of details. The use of global context information and fractionally-strided convolutional layers results in better estimation quality. Additionally, the use of  
local context and minimization over a weighted combination of $L_{A}$ and $L_E$ further improves the quality and reduces the estimation error.

\subsection{Evaluations and comparisons}
In this section, the results of the proposed method are compared against recent state-of-the-art methods on three challenging datasets. 

\noindent {\bf{ShanghaiTech.}} The proposed method is evaluated against four recent approaches: Zhang \etal \cite{zhang2015cross}, MCNN  \cite{zhang2016single}, Cascaded-MTL \cite{sindagi2017cnnbased} and Switching-CNN \cite{sam2017switching} on Part A and Part B of the ShanghaiTech dataset are shown in Table \ref{tab:resultsshanghaitech}. The authors in \cite{zhang2015cross} proposed a switchable learning function where they learned their network by alternatively training on two objective functions: crowd count and density estimation. They made use of perspective maps for appropriate ground truth density maps.  
In another approach, Zhang \etal  \cite{zhang2016single} proposed a multi-column convolutional network (MCNN) to address scale issues and a sophisticated ground truth density map generation technique. Instead of using the responses of all the columns, Sam \etal \cite{sam2017switching} proposed a switching-CNN classifier that chooses the optimal regressor. Sindagi \etal \cite{sindagi2017cnnbased} incorporate high-level prior in the form of crowd density levels and perform a cascaded multi-task learning of estimating prior and density map. It can be observed from Table \ref{tab:resultsshanghaitech}, that the proposed method is able to achieve superior results as compared to the other methods, which highlights the importance of contextual processing in our framework. 

\begin{table}[htp!]
\centering
\resizebox{0.77\linewidth}{!}{
\begin{tabular}{|l|c|c|c|c|}
\hline
                & \multicolumn{2}{c|}{Part A} & \multicolumn{2}{c|}{Part B} \\ \hline
Method          & MAE          & MSE          & MAE          & MSE          \\ \hline\hline
Zhang \etal \cite{zhang2015cross}    & 181.8        & 277.7        & 32.0         & 49.8         \\ \hline
MCNN \cite{zhang2016single}           & 110.2        & 173.2        & 26.4         & 41.3         \\ \hline
Cascaded-MTL \cite{sindagi2017cnnbased}           & 101.3        & 152.4        & 20.0         & 31.1         \\ \hline
Switching-CNN \cite{sam2017switching}           & 90.4        & 135.0        & 21.6         & 33.4         \\ \hline
CP-CNN (ours) & \textbf{73.6}        & \textbf{106.4}        & \textbf{20.1}         & \textbf{30.1}         \\ \hline
\end{tabular}
}
\vskip -8pt\caption{Estimation errors on the ShanghaiTech dataset.}
\label{tab:resultsshanghaitech}
\end{table}

\noindent {\bf{WorldExpo'10.}}  The WorldExpo'10 dataset was introduced by Zhang \etal \cite{zhang2015cross} and it contains 3,980 annotated frames from 1,132 video sequences captured by 108 surveillance cameras. The frames are divided into training and test sets. The training set contains 3,380 frames and the test set contains 600 frames from five different scenes with 120 frames per scene. They also provided Region of Interest (ROI) map for each of the five scenes. For a fair comparison, perspective maps were used to generate the ground truth maps similar to the work of \cite{zhang2015cross}. Also, similar to \cite{zhang2015cross}, ROI maps are considered for post processing the output density map generated by the network. 

The proposed method  is evaluated against five recent state-of-the-art approaches: Chen \etal \cite{chen2013cumulative}, Zhang \etal \cite{zhang2015cross}, MCNN \cite{zhang2016single}, Shang \etal \cite{skaug2016end} and Switching-CNN \cite{sam2017switching} is presented in Table \ref{tab:resultsworldexpo}. The authors in \cite{chen2013cumulative} introduced cumulative attributive concept for learning a regression model for crowd density and age estimation. Shang \etal \cite{skaug2016end} proposed an end-to-end CNN architecture consisting of three parts: pre-trained GoogLeNet model for feature generation, long short term memory (LSTM) decoders for local count and fully connected layers for the final count. It can be observed from Table \ref{tab:resultsworldexpo} that the proposed method outperforms existing approaches on an average while achieving comparable performance in individual scene estimations. 

\begin{table}[htp!]
\centering
\resizebox{0.99\linewidth}{!}{
\begin{tabular}{|l|c|c|c|c|c|c|}
\hline
Method & Scene1 & Scene2 & Scene3 & Scene4 & Scene5 & Avgerage\\
\hline\hline
Chen \etal \cite{chen2013cumulative} &\textbf{2.1} & 55.9 & \textbf{9.6} & 11.3 & \textbf{3.4} & 16.5\\
\hline
Zhang \etal \cite{zhang2015cross} &9.8 & \textbf{14.1} & 14.3 & 22.2 & 3.7 & 12.9\\
\hline
MCNN \cite{zhang2016single}& 3.4 & 20.6 & 12.9 & 13.0 & 8.1 & 11.6\\
\hline
Shang \etal \cite{skaug2016end}& 7.8 & 15.4 & 14.9 & 11.8 & 5.8 & 11.7\\
\hline
Switching-CNN \cite{sam2017switching}& 4.4 & 15.7 & 10.0 & 11.0 & 5.9 & 9.4\\ 
\hline
CP-CNN (ours) & 2.9 & 14.7 & 10.5 & \textbf{10.4} & 5.8 & \textbf{8.86}\\
\hline
\end{tabular}
}
\vskip -8pt \caption{Average estimation errors on the WorldExpo'10 dataset. }
 \label{tab:resultsworldexpo}
\end{table}

\noindent {\bf{UCF\textunderscore CC\textunderscore 50.}}  The UCF\textunderscore CC\textunderscore 50 is an extremely challenging dataset introduced by Idrees \etal \cite{idrees2013multi}. The dataset contains 50 annotated images of different resolutions and aspect ratios crawled from the internet. There is a large variation in densities across images. Following the standard protocol discussed in \cite{idrees2013multi}, a 5-fold cross-validation was performed for evaluating the proposed method. Results are compared with seven recent approaches: Idrees \etal \cite{idrees2013multi}, Zhang \etal \cite{zhang2015cross}, MCNN \cite{zhang2016single}, Onoro \etal \cite{onoro2016towards}, Walach \etal \cite{walach2016learning}, Cascaded-MTL \cite{sindagi2017cnnbased} and Switching-CNN  \cite{sam2017switching}. The authors in \cite{idrees2013multi} proposed to combine information from multiple sources such as head detections, Fourier analysis and texture features (SIFT). Onoro \etal in \cite{onoro2016towards} proposed a scale-aware CNN to learn a multi-scale non-linear regression model using a pyramid of image patches extracted at multiple scales. Walach \etal \cite{walach2016learning} proposed a layered approach of learning CNNs for crowd counting by iteratively adding CNNs where every new CNN is trained on residual error of the previous layer. It can be observed from Table \ref{tab:resultsucf} that our network achieves the lowest MAE and MSE count errors.  This experiment clearly shows the significance of using context especially in images with widely varying densities.

\begin{table}[htp!]
\centering
\resizebox{0.65\linewidth}{!}{
\begin{tabular}{|l|c|c|}
\hline
Method & MAE & MSE \\
\hline\hline
Idrees \etal \cite{idrees2013multi}& 419.5 & 541.6 \\
\hline
Zhang \etal \cite{zhang2015cross}& 467.0 & 498.5 \\
\hline
MCNN \cite{zhang2016single} & 377.6 & 509.1\\
\hline
Onoro \etal \cite{onoro2016towards}  Hydra-2s& 333.7 & 425.2 \\
\hline
Onoro \etal \cite{onoro2016towards} Hydra-3s& 465.7 & 371.8 \\
\hline
Walach \etal \cite{walach2016learning} & 364.4 & 341.4 \\
\hline
Cascaded-MTL \cite{sindagi2017cnnbased} & 322.8.4 & 341.4 \\
\hline
Switching-CNN \cite{sam2017switching} & 318.1 & 439.2 \\
\hline
CP-CNN (ours) & \textbf{295.8} & \textbf{320.9} \\
\hline
\end{tabular}
}
\vskip -8pt\caption{Estimation errors on the UCF\textunderscore CC\textunderscore 50 dataset.}
\label{tab:resultsucf}
\end{table}

\section{Conclusion}
\label{sec:conclusion}
We presented contextual pyramid of CNNs for incorporating global and local contextual information in an image to generate high-quality crowd density maps and lower count estimation errors. The global and local contexts are obtained by learning to classify the input images and its patches into various density levels. This context information is then fused with the output of a multi-column DME by a Fusion-CNN. In contrast to the existing methods, this work focuses on generating better quality density maps in addition to achieving lower count errors. In this attempt, the Fusion-CNN is constructed with fractionally-strided convolutional layers and it is trained along with the DME in an end-to-end fashion by optimizing a weighted combination of adversarial loss and pixel-wise Euclidean loss.    Extensive experiments performed on challenging datasets and comparison with recent state-of-the-art approaches demonstrated the significant improvements achieved by the proposed method. 


\section*{Acknowledgement}
This work was supported by US Office of Naval Research (ONR) Grant
YIP N00014-16-1-3134.

{\small
\bibliographystyle{ieee}
\bibliography{egbib}
}

\clearpage
\onecolumn
\section*{Appendix}

\begin{figure*}[ht!]
	\begin{center}
		\begin{subfigure}[t]{0.33\textwidth}
			\includegraphics[width=\textwidth]{figures/s/sa/IMG_140.jpg}
			\includegraphics[width=\textwidth]{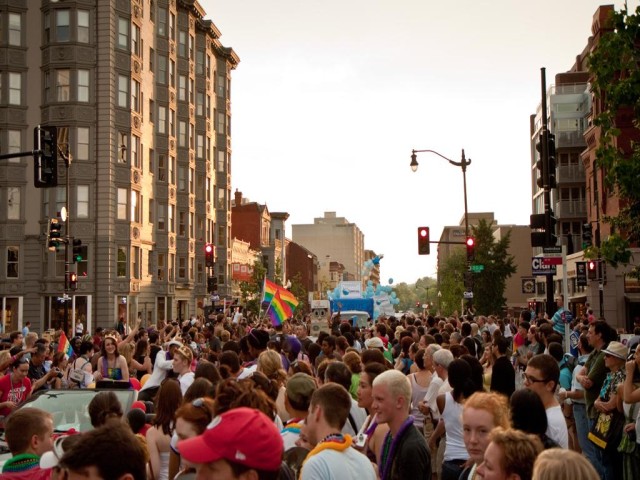}
			\includegraphics[width=\textwidth]{figures/s/sa/IMG_146.jpg}
			\includegraphics[width=\textwidth]{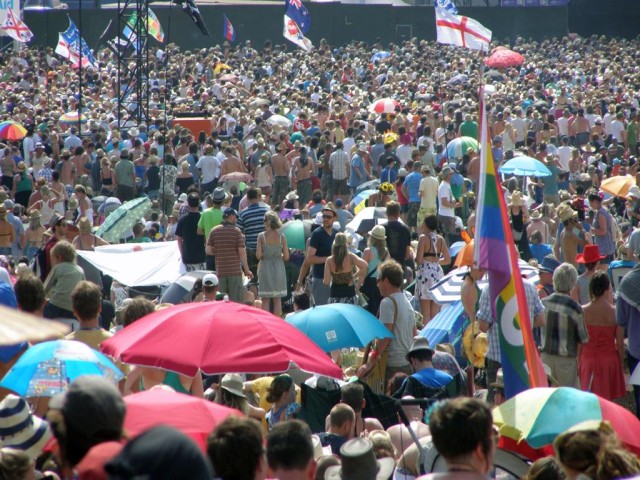}
		\end{subfigure}
		\begin{subfigure}[t]{0.33\textwidth}
			\includegraphics[width=\textwidth]{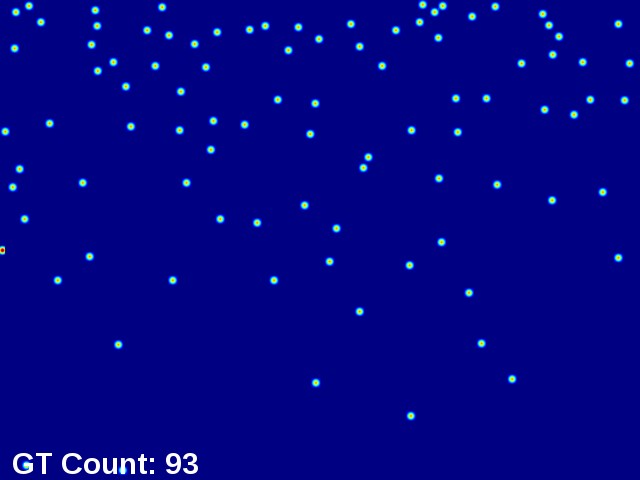}
			\includegraphics[width=\textwidth]{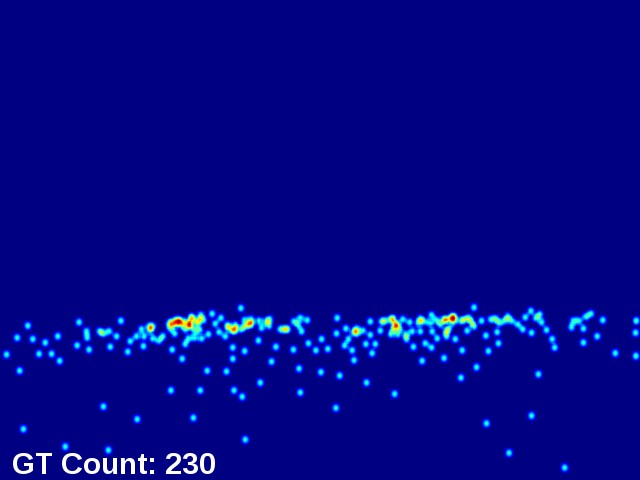}
			\includegraphics[width=\textwidth]{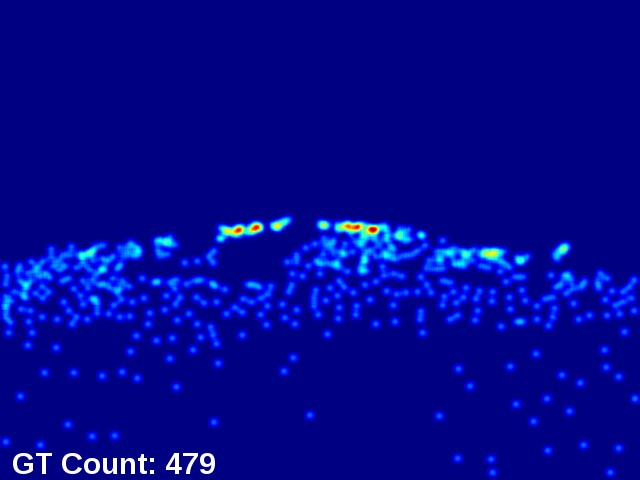}
			\includegraphics[width=\textwidth]{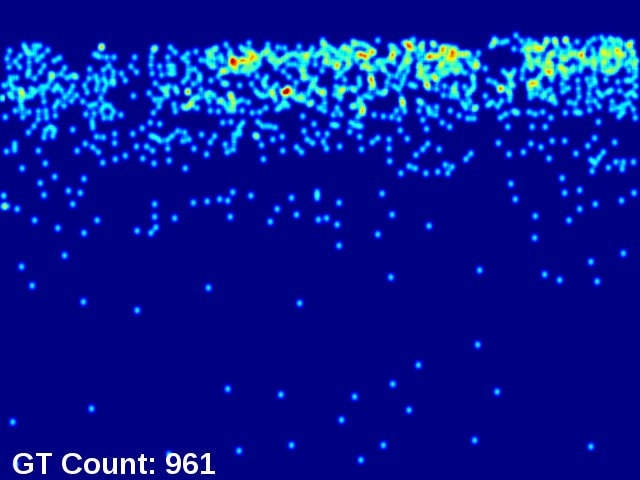}
		\end{subfigure}
		\begin{subfigure}[t]{0.33\textwidth}
			\includegraphics[width=\textwidth]{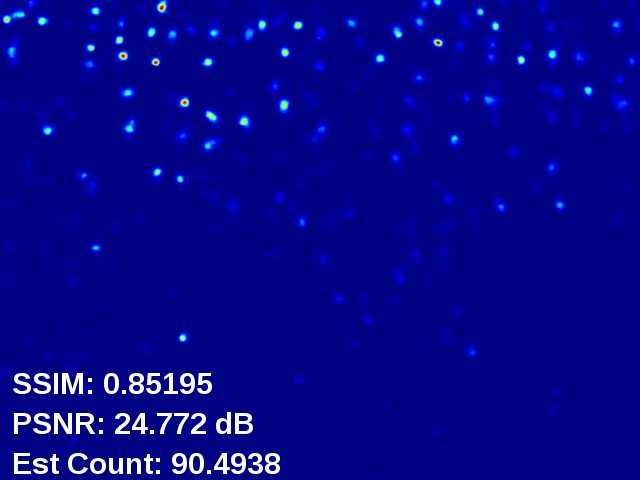}
			\includegraphics[width=\textwidth]{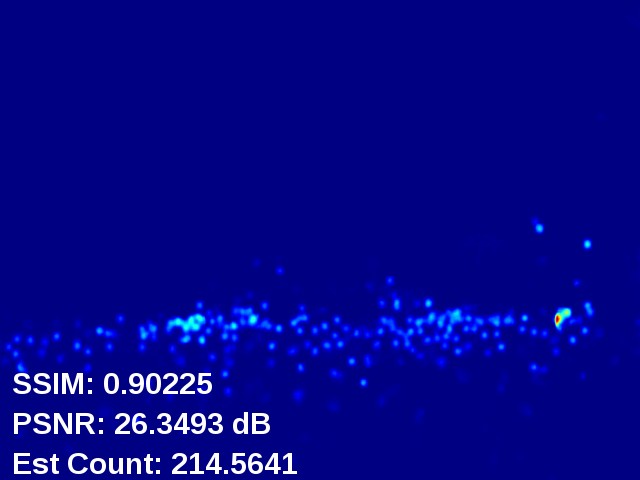}
			\includegraphics[width=\textwidth]{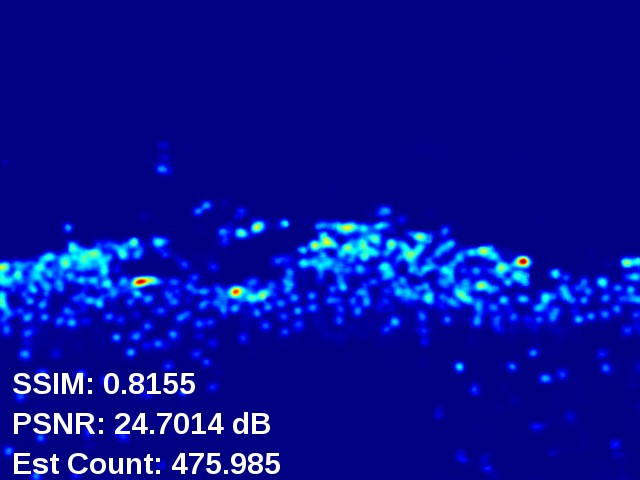}
			\includegraphics[width=\textwidth]{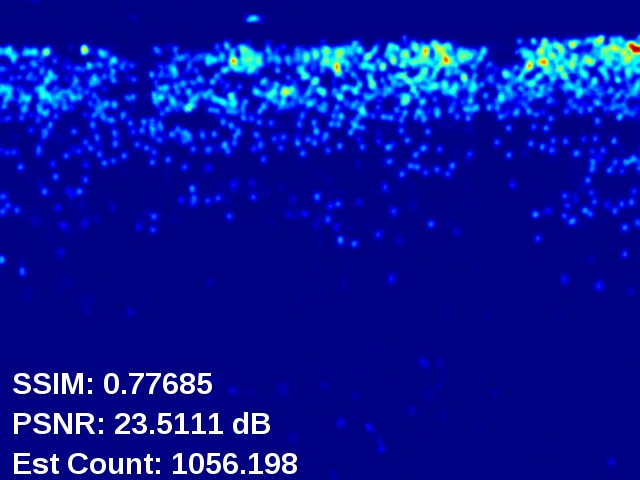}
		\end{subfigure}
	\end{center}
	\vskip -16pt\caption{Results of the proposed CP-CNN method on Shanghai Tech Part A dataset \cite{zhang2016single}. \textit{Left column}: Input images. \textit{Middle column}: Ground truth density maps. \textit{Right column}: Estimated density maps.}
	\label{fig:sa}
\end{figure*}

\begin{figure*}[ht!]
	\begin{center}
		\begin{subfigure}[t]{0.33\textwidth}
			\includegraphics[width=\textwidth]{figures/s/sb/IMG_14.jpg}
			\includegraphics[width=\textwidth]{figures/s/sb/IMG_21.jpg}
			\includegraphics[width=\textwidth]{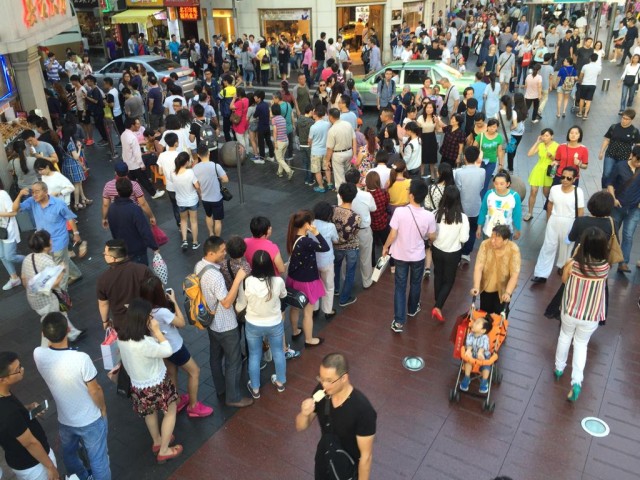}
			\includegraphics[width=\textwidth]{figures/s/sb/IMG_281.jpg}
			\includegraphics[width=\textwidth]{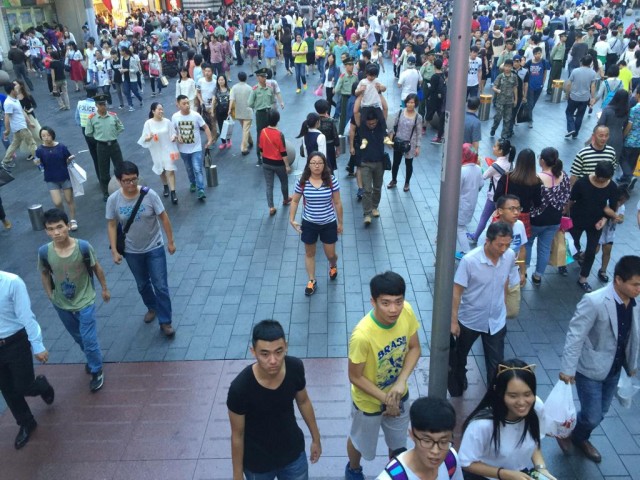}
		\end{subfigure}
		\begin{subfigure}[t]{0.33\textwidth}
			\includegraphics[width=\textwidth]{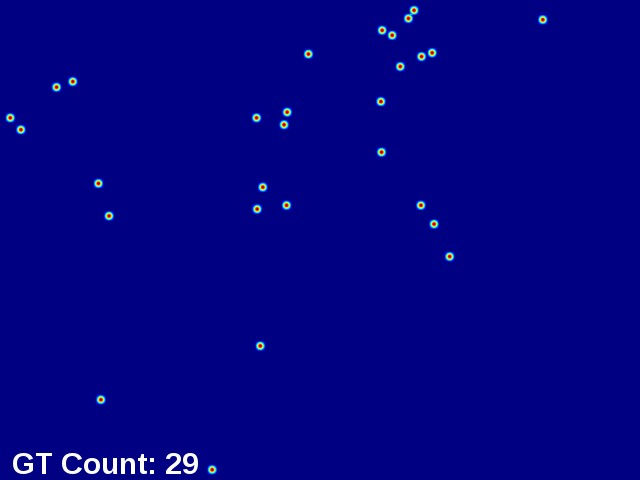}
			\includegraphics[width=\textwidth]{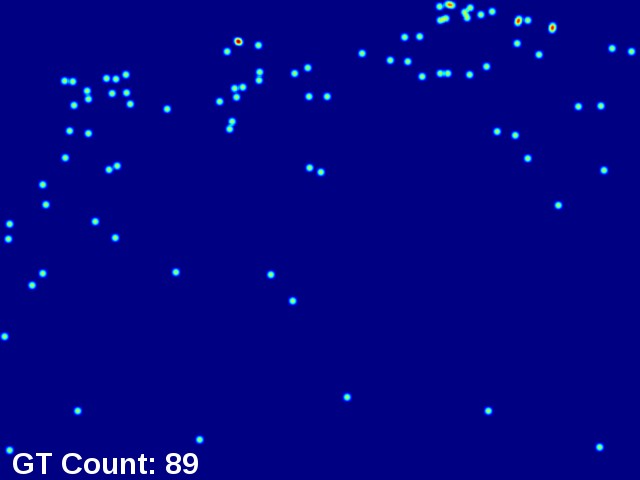}
			\includegraphics[width=\textwidth]{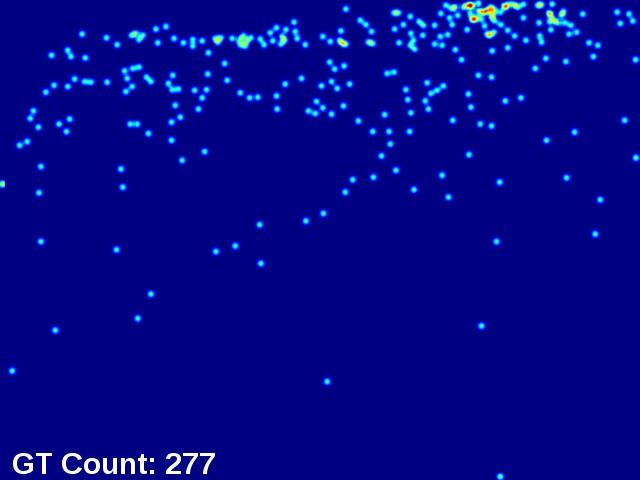}
			\includegraphics[width=\textwidth]{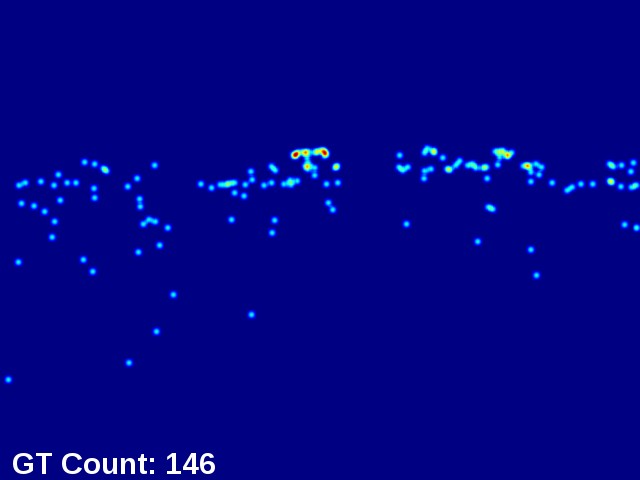}
			\includegraphics[width=\textwidth]{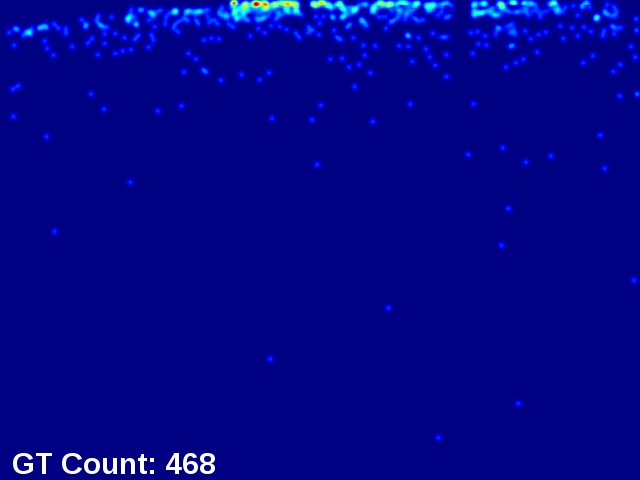}
		\end{subfigure}
		\begin{subfigure}[t]{0.33\textwidth}
			\includegraphics[width=\textwidth]{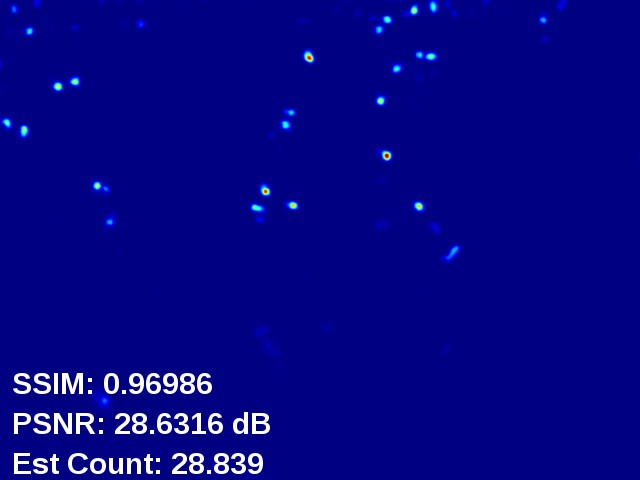}
			\includegraphics[width=\textwidth]{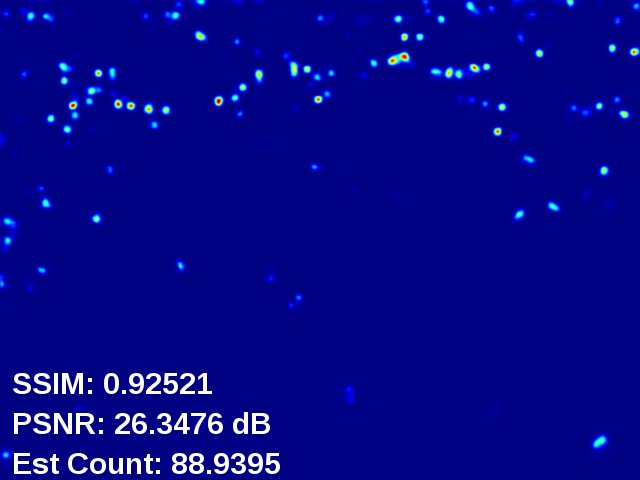}
			\includegraphics[width=\textwidth]{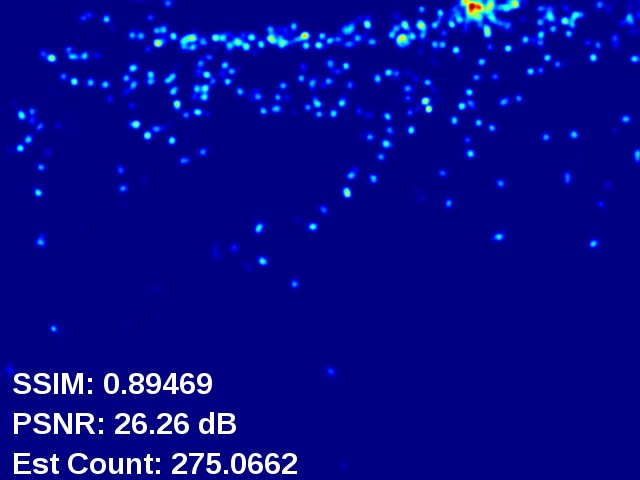}
			\includegraphics[width=\textwidth]{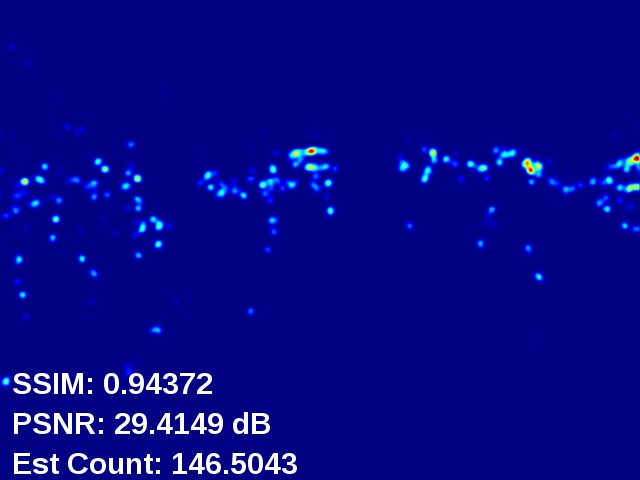}
			\includegraphics[width=\textwidth]{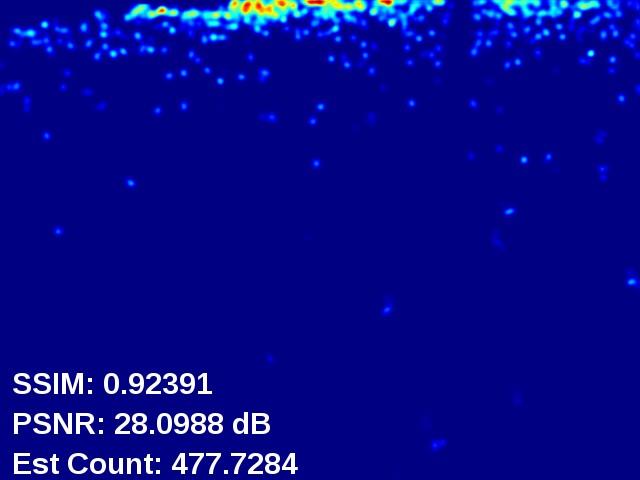}
		\end{subfigure}
	\end{center}
	\vskip -16pt\caption{Results of the proposed CP-CNN method on Shanghai Tech Part B dataset \cite{zhang2016single}. \textit{Left column}: Input images. \textit{Middle column}: Ground truth density maps. \textit{Right column}: Estimated density maps.}
	\label{fig:sb}
\end{figure*}

\begin{figure*}[ht!]
	\begin{center}
		\begin{subfigure}[t]{0.33\textwidth}
			\includegraphics[width=\textwidth]{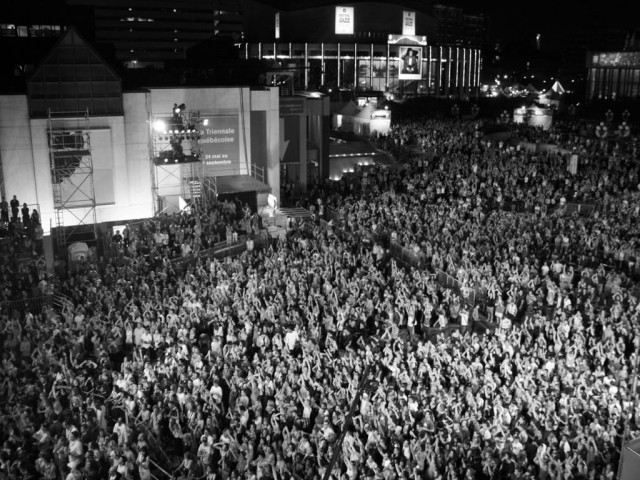}
			\includegraphics[width=\textwidth]{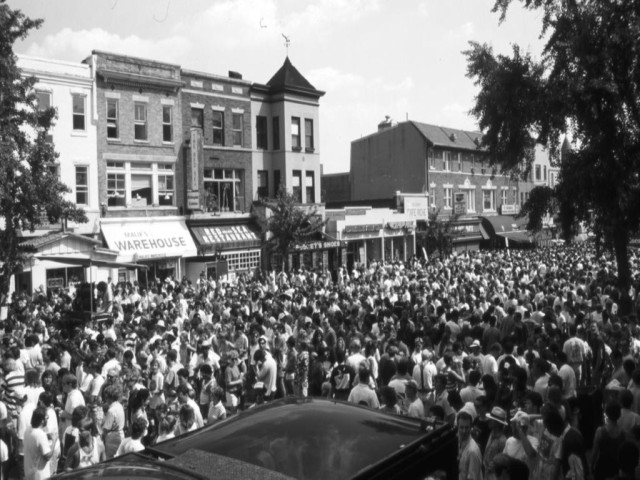}
			\includegraphics[width=\textwidth]{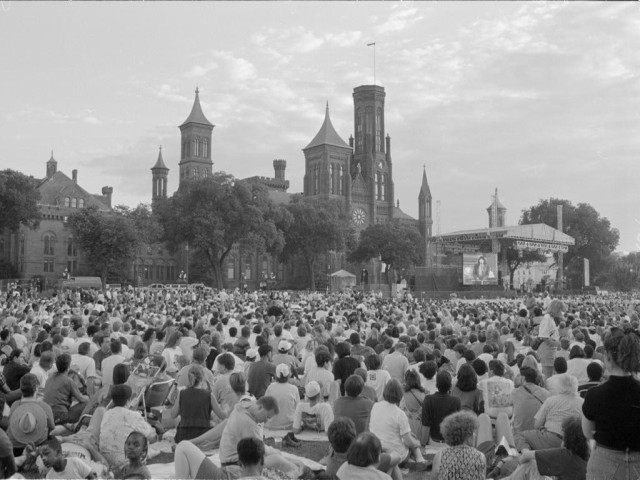}
			\includegraphics[width=\textwidth]{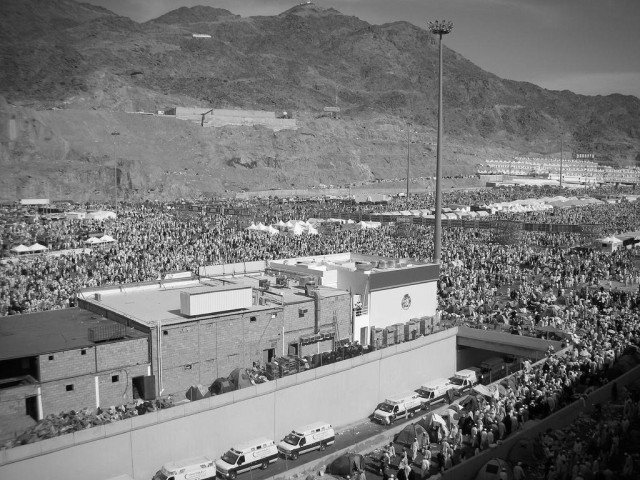}
			\includegraphics[width=\textwidth]{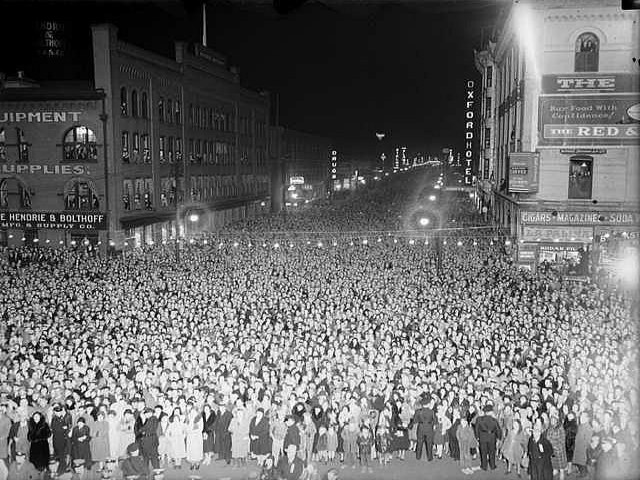}
		\end{subfigure}
		\begin{subfigure}[t]{0.33\textwidth}
			\includegraphics[width=\textwidth]{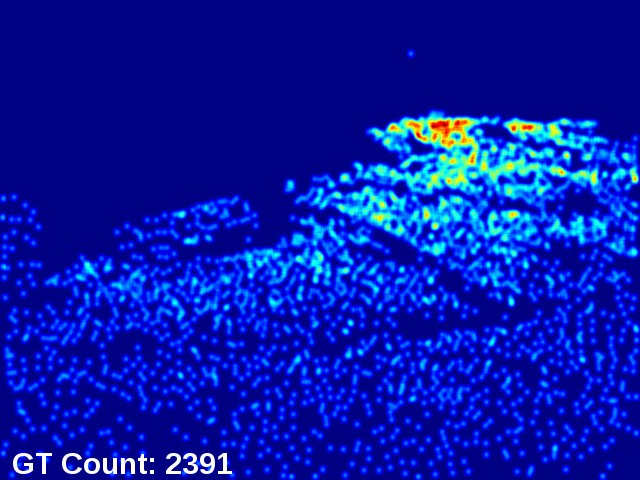}
			\includegraphics[width=\textwidth]{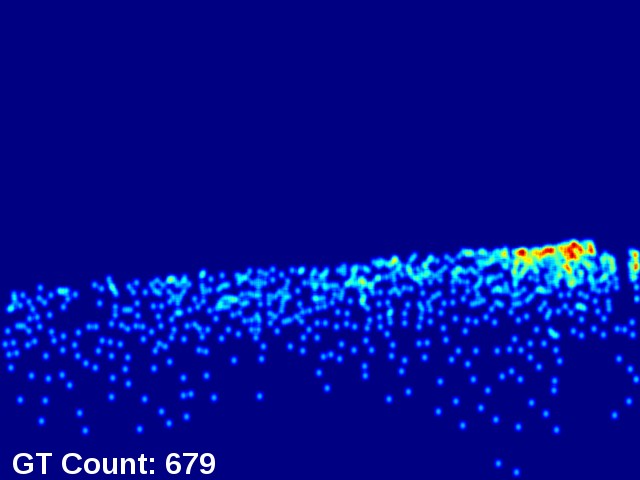}
			\includegraphics[width=\textwidth]{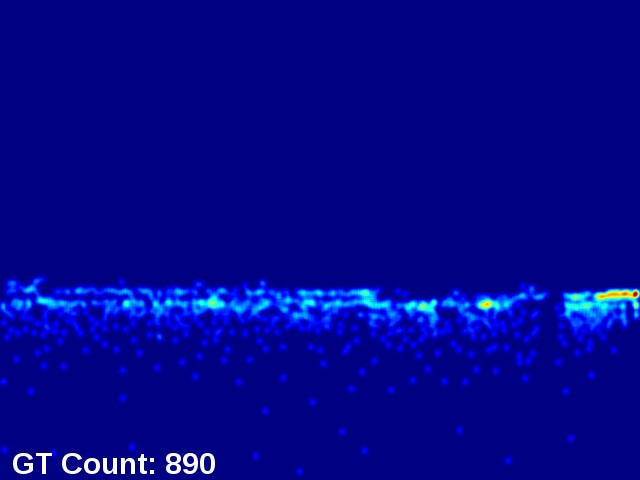}
			\includegraphics[width=\textwidth]{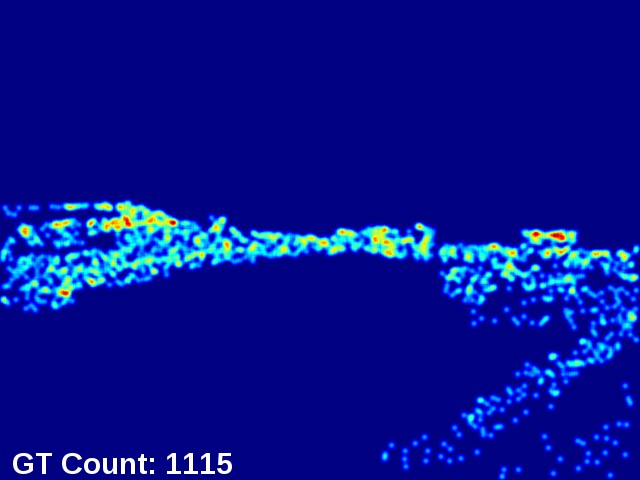}
			\includegraphics[width=\textwidth]{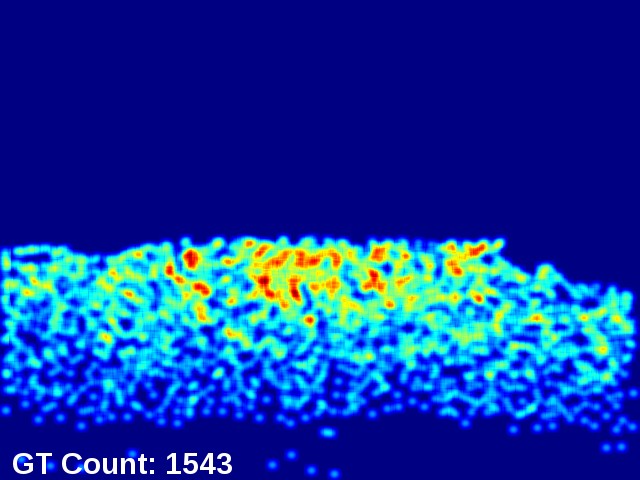}
		\end{subfigure}
		\begin{subfigure}[t]{0.33\textwidth}
			\includegraphics[width=\textwidth]{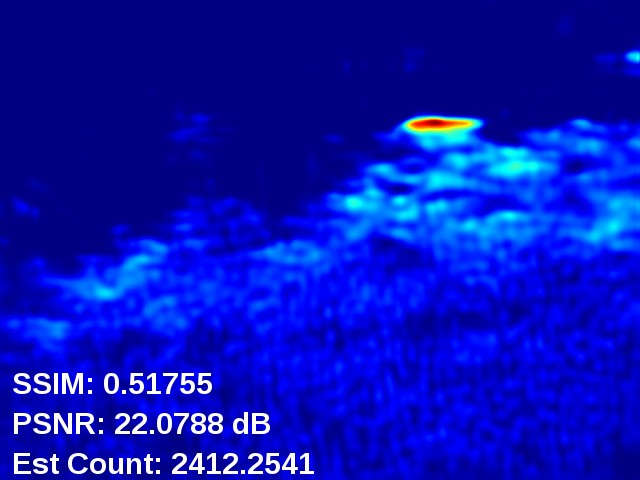}
			\includegraphics[width=\textwidth]{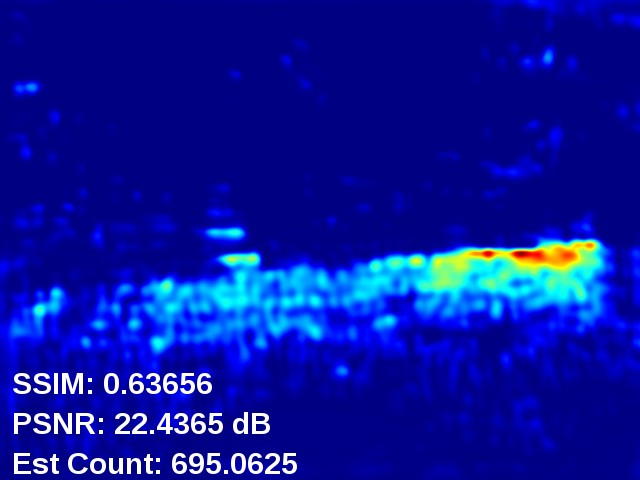}
			\includegraphics[width=\textwidth]{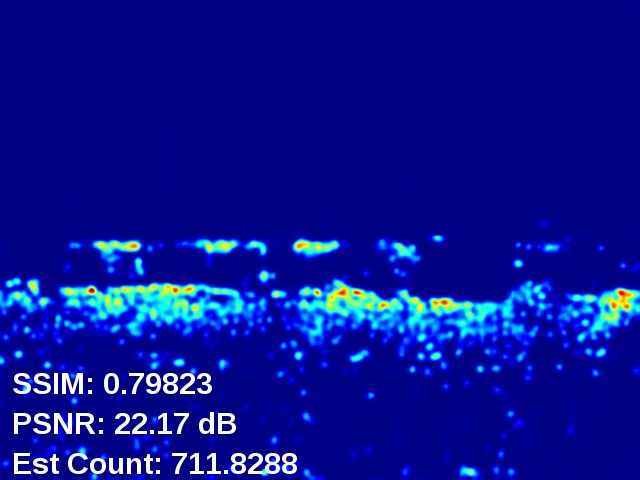}
			\includegraphics[width=\textwidth]{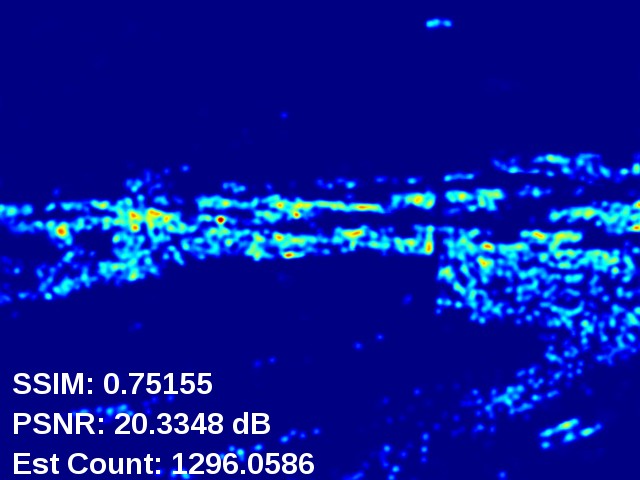}
			\includegraphics[width=\textwidth]{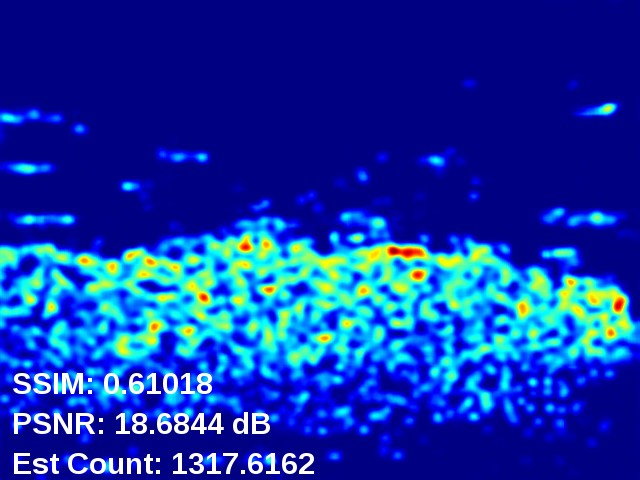}
		\end{subfigure}
	\end{center}
	\vskip -16pt\caption{Results of the proposed CP-CNN method on UCF\textunderscore CC\textunderscore 50 dataset \cite{idrees2013multi}. \textit{Left column}: Input images. \textit{Middle column}: Ground truth density maps. \textit{Right column}: Estimated density maps.}
	\label{fig:u}
\end{figure*}

\begin{figure*}[ht!]
	\begin{center}
		\begin{subfigure}[t]{0.33\textwidth}
			\includegraphics[width=\textwidth]{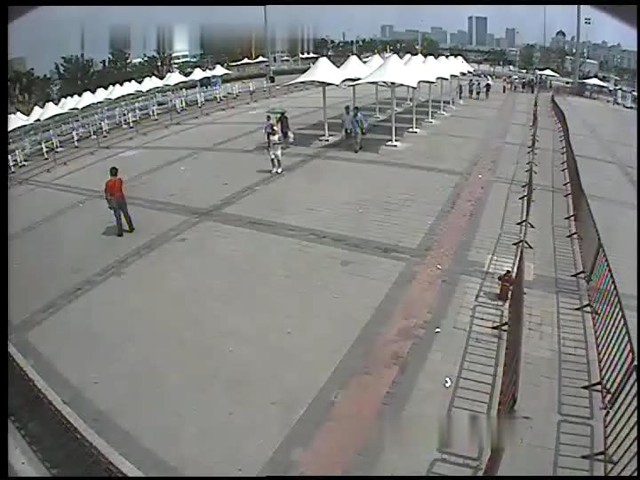}
			\includegraphics[width=\textwidth]{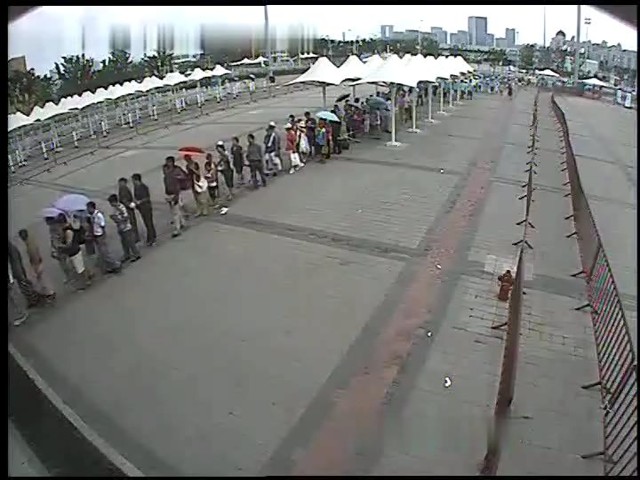}
			\includegraphics[width=\textwidth]{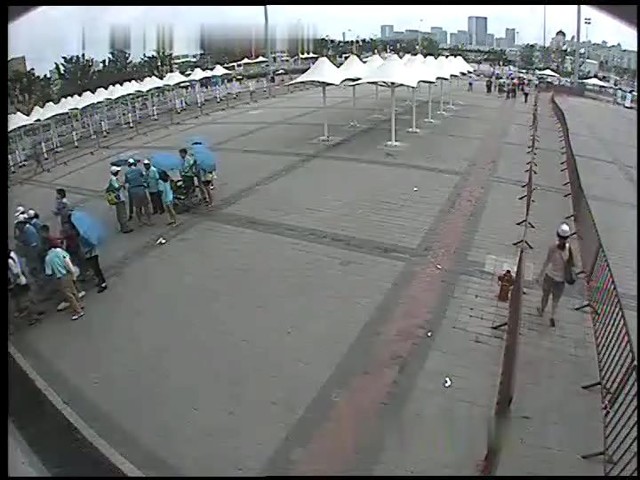}
			\includegraphics[width=\textwidth]{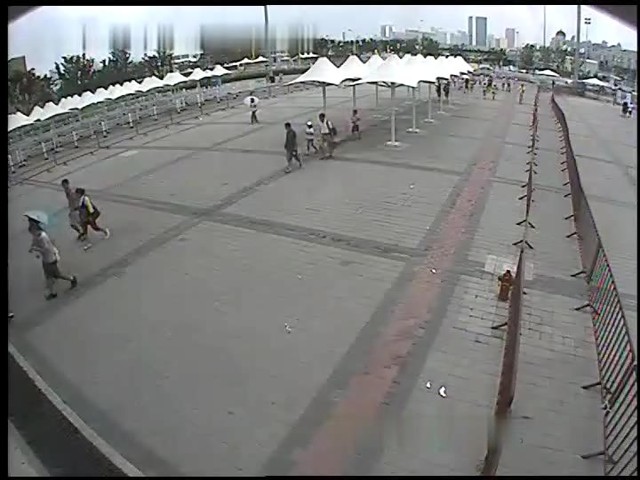}
			\includegraphics[width=\textwidth]{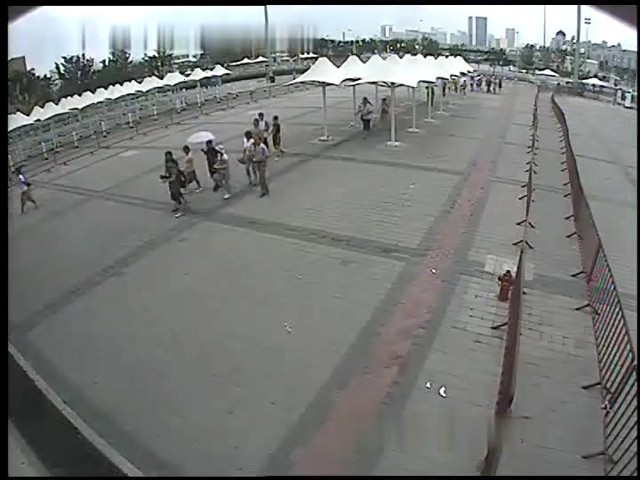}
		\end{subfigure}
		\begin{subfigure}[t]{0.33\textwidth}
			\includegraphics[width=\textwidth]{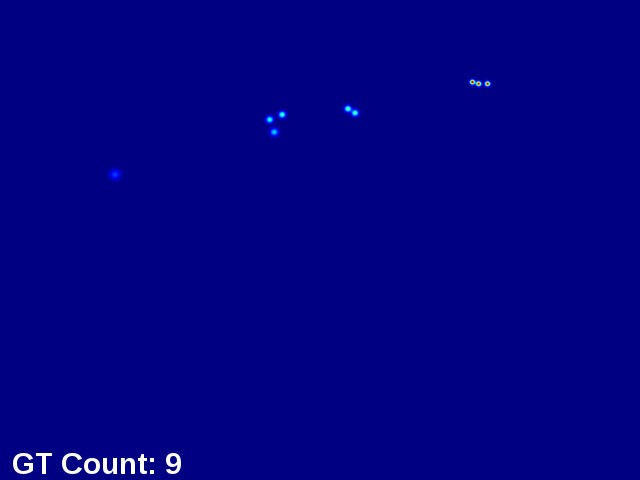}
			\includegraphics[width=\textwidth]{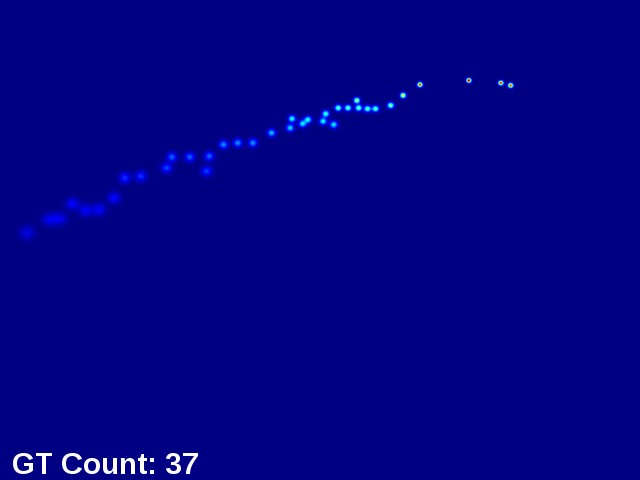}
			\includegraphics[width=\textwidth]{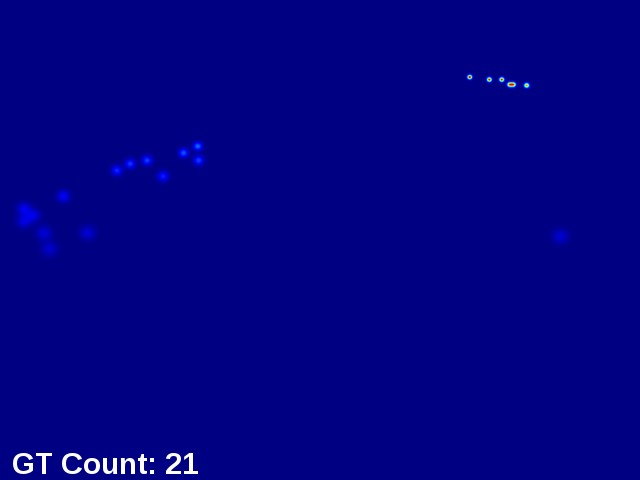}
			\includegraphics[width=\textwidth]{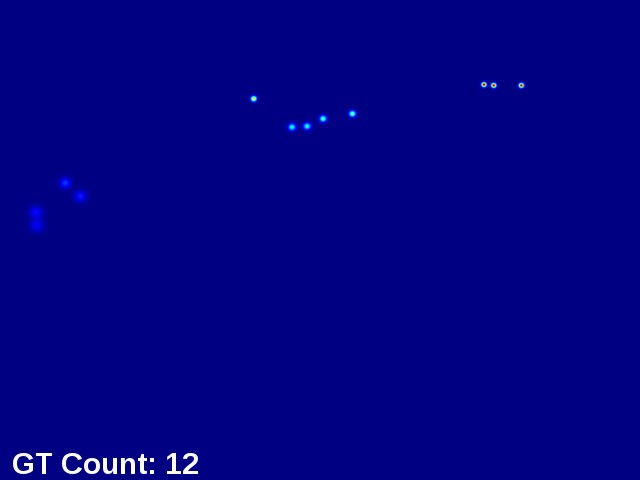}
			\includegraphics[width=\textwidth]{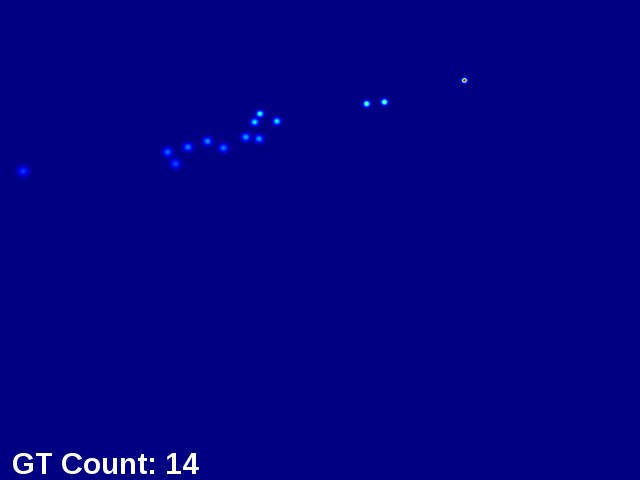}
		\end{subfigure}
		\begin{subfigure}[t]{0.33\textwidth}
			\includegraphics[width=\textwidth]{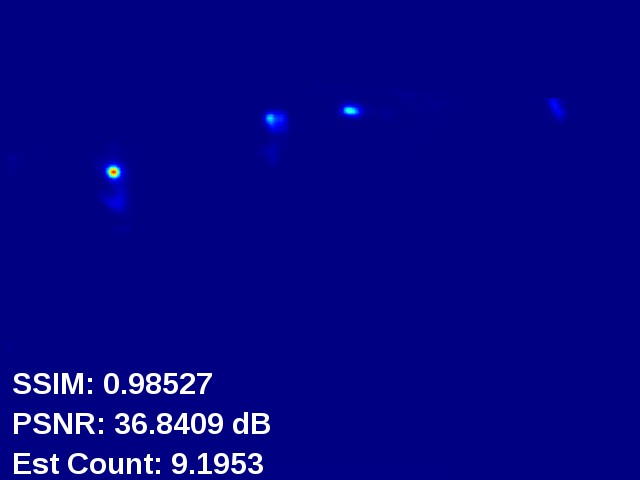}
			\includegraphics[width=\textwidth]{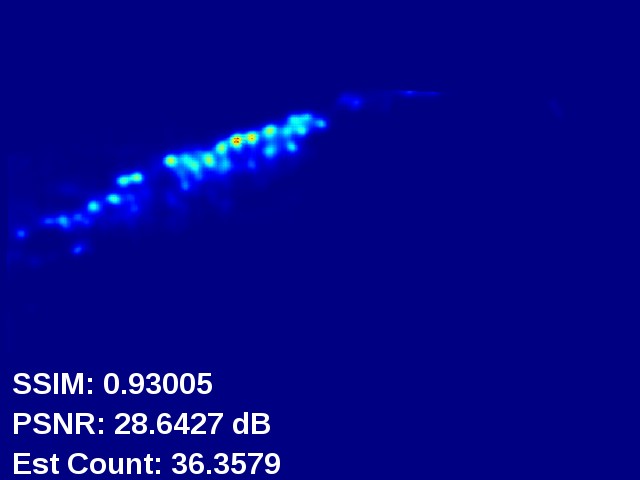}
			\includegraphics[width=\textwidth]{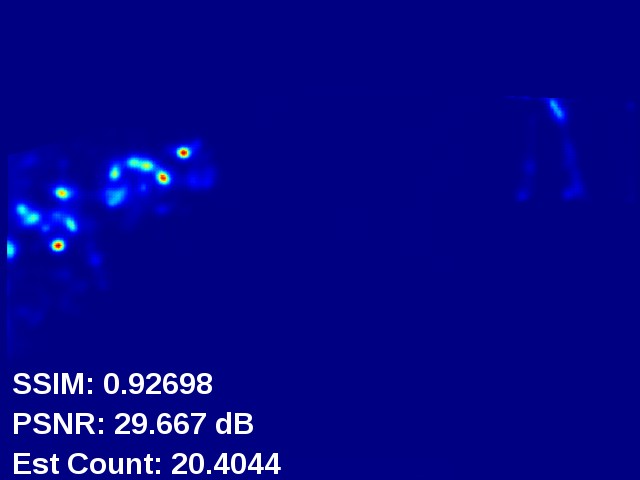}
			\includegraphics[width=\textwidth]{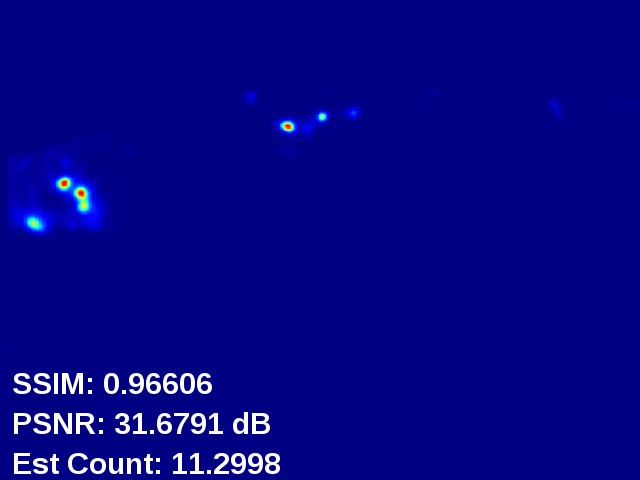}
			\includegraphics[width=\textwidth]{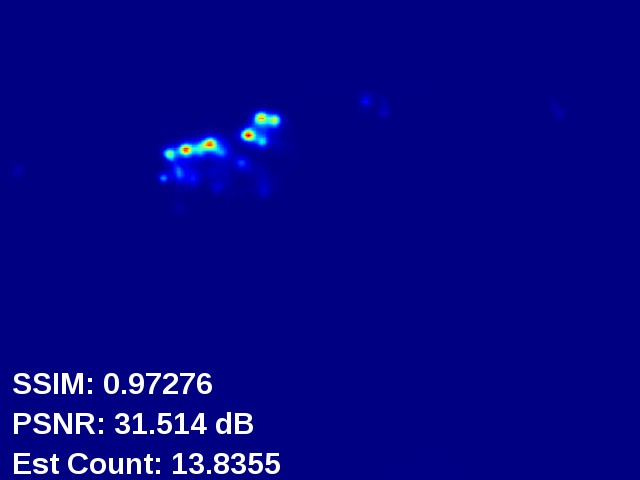}
		\end{subfigure}
	\end{center}
	\vskip -16pt\caption{Results of the proposed CP-CNN method on WorldExpo '10 dataset \cite{zhang2015cross}. \textit{Left column}: Input images. \textit{Middle column}: Ground truth density maps. \textit{Right column}: Estimated density maps.}
	\label{fig:w}
\end{figure*}

This section contains some additional results of the proposed method for the three datasets (Shanghai Tech \cite{zhang2016single}, UCF\textunderscore CC\textunderscore 50 dataset \cite{idrees2013multi} and WorldExpo '10 \cite{zhang2015cross}) on which the evaluations were performed. Results on sample images from these datasets are shown in Fig. \ref{fig:sa} to Fig. \ref{fig:w}. Sample images were chosen carefully to be representative of various density levels present in the respective datasets.

\end{document}